\documentclass[conference]{IEEEtran}
\IEEEoverridecommandlockouts
\usepackage{microtype}
\usepackage{graphicx}
\usepackage{subfigure}
\usepackage{booktabs} 
\usepackage{paralist}
\usepackage{amssymb}
\usepackage{amsmath}
\usepackage{empheq}
\usepackage{mdframed}
\usepackage{xspace}
\usepackage[utf8]{inputenc}
\usepackage[english]{babel}
\usepackage{amsthm} 
\usepackage{bbding}
\usepackage{diagbox}
\usepackage{color, colortbl}
\usepackage{cellspace}
\usepackage{multirow}
\usepackage{enumitem}

\usepackage{hyperref}

\def\BibTeX{{\rm B\kern-.05em{\sc i\kern-.025em b}\kern-.08em
    T\kern-.1667em\lower.7ex\hbox{E}\kern-.125emX}}

\newtheorem{definition}{Definition}

\newcommand{\hide}[1]{}

\newcommand{\BFSERIES}{\fontseries{b}\selectfont}

\newcommand{\fnddag}{\textsuperscript{$\ddagger$}}
\newcommand{\fndag}{\textsuperscript{$\dagger$}}

\newcommand\Tstrut{\rule{0pt}{2.6ex}}         
\newcommand\Bstrut{\rule[-0.9ex]{0pt}{0pt}}   

\newcommand{\bit}{\begin{compactitem}}
\newcommand{\eit}{\end{compactitem}}
\newcommand{\ben}{\begin{compactenum}}
\newcommand{\een}{\end{compactenum}}

\newcommand{\beq}{\begin{equation}}
\newcommand{\eeq}{\end{equation}}

\newcommand{\hy}{{\mathbf{\widehat{y}}}}

\newcommand{\bh}{\mathbf{h}}
\newcommand{\be}{\mathbf{e}}
\newcommand{\hdos}{\mathbf{h}^{\text{DOS}}}
\newcommand{\hldos}{\mathbf{h}^{\text{LDOS}}}
\newcommand{\hcldos}{\mathbf{h}^{\text{cLDOS}}}

\newcommand{\R}{\mathbb{R}} 
\newcommand{\I}{\mathbb{I}}

\newcommand{\frf}{{\sc FRF}\xspace}

\newcommand{\bc}{\mathbf{c}}
\newcommand{\bv}{\mathbf{v}}
\newcommand{\bvp}{\mathbf{v}^\prime}
\newcommand{\bu}{\mathbf{u}}
\newcommand{\x}{\mathbf{x}}
\newcommand{\z}{\mathbf{z}}
\newcommand{\X}{\mathbf{X}}
\newcommand{\W}{\mathbf{W}}
\newcommand{\tW}{{\mathbf{\widetilde{W}}}}
\newcommand{\tL}{{\mathbf{\widetilde{L}}}}
\newcommand{\D}{\mathbf{D}}
\newcommand{\U}{\mathbf{U}}
\newcommand{\bP}{\mathbf{P}}

\newcommand{\Lam}{\boldsymbol{\Lambda}}
\newcommand{\bS}{\mathbf{S}}
\newcommand{\bphi}{\boldsymbol{\phi}}

\newcommand{\method}{{\sc A-DOGE}\xspace}
\newcommand{\doge}{{\sc DOGE}\xspace}

\newcommand{\mX}{\mathcal{X}}
\newcommand{\mV}{\mathcal{V}}
\newcommand{\mE}{\mathcal{E}}

\newcommand{\redditb}{\texttt{REDDIT-B}\xspace}
\newcommand{\reddit}{\texttt{REDDIT-5K}\xspace}
\newcommand{\collab}{\texttt{COLLAB}\xspace}
\newcommand{\imdbb}{\texttt{IMDB-BIN}\xspace}
\newcommand{\imdb}{\texttt{IMDB-MUL}\xspace}
\newcommand{\dd}{\texttt{DD}\xspace}
\newcommand{\protein}{\texttt{PROTEINS}\xspace}
\newcommand{\aids}{\texttt{AIDS}\xspace}

\newcommand{\congress}{\texttt{Congress}\xspace}
\newcommand{\congressl}{\texttt{Congress-l}\xspace}
\newcommand{\mig}{\texttt{MIG}\xspace}
\newcommand{\band}{\texttt{BandPass}\xspace}

\newcommand{\states}{\texttt{BorderStates}\xspace}
\newcommand{\face}{\texttt{Facebook100}\xspace}

\newcommand{\sbt}{\,\begin{picture}(-1,1)(-1,-3)\circle*{3}\end{picture}\ }

\newcommand{\fgsd}{\textsc{FGSD}\xspace}
\newcommand{\netlsd}{\textsc{NetLSD}\xspace}
\newcommand{\gvec}{\textsc{g2vec}\xspace}

\newcommand{\wl}{\textsc{WL}\xspace}
\newcommand{\wloa}{\textsc{WL-OA}\xspace}	
\newcommand{\pk}{\textsc{PK}\xspace}
\newcommand{\retgk}{\textsc{RetGK}\xspace}
\newcommand{\sage}{\textsc{SAGE}\xspace}
\newcommand{\dos}{\textsc{DOSGK}\xspace}

\newcommand{\gcn}{\textsc{GCN}\xspace}
\newcommand{\gin}{\textsc{GIN}\xspace}
\newcommand{\cheb}{\textsc{ChebNet}\xspace}
\newcommand{\caley}{\textsc{CaleyNet}\xspace}

\newtheorem{problem}{Problem}

\long\def\ignore#1{}

\begin{document}

\title{{\fontsize{23}{23}\selectfont Fast Attributed Graph Embedding via Density of States}}

\author{\IEEEauthorblockN{Saurabh Sawlani}
\IEEEauthorblockA{
\textit{Carnegie Mellon University}\\
saurabh.sawlani@gmail.com}
\and
\IEEEauthorblockN{Lingxiao Zhao}
\IEEEauthorblockA{
\textit{Carnegie Mellon University}\\
lingxia1@andrew.cmu.edu }
\and
\IEEEauthorblockN{Leman Akoglu}
\IEEEauthorblockA{
\textit{Carnegie Mellon University}\\
lakoglu@andrew.cmu.edu}
}

\maketitle

\begin{abstract}
Given a node-attributed graph, how can we efficiently represent it with few numerical features that expressively reflect its topology and attribute information?
We propose \method, for {\sc A}ttributed {\sc DO}S-based {\sc G}raph {\sc E}mbedding, based on density of states (DOS, a.k.a. spectral density) to tackle this problem.
\method is designed to fulfill a long desiderata of desirable characteristics. 
Most notably, it capitalizes on efficient approximation algorithms for DOS, that we extend to blend in node labels and attributes for the first time, making it {fast and scalable} for large attributed graphs and graph databases.	 
Being based on the entire eigenspectrum of a graph, \method can capture structural and attribute properties at {multiple (``glocal'') scales}.
Moreover, it is {unsupervised}
 (i.e. agnostic to any specific objective) and lends itself to various interpretations, 
which makes it is suitable for exploratory graph mining tasks.
Finally,
it {processes each graph independent of others}, making it amenable for streaming settings as well as parallelization.
Through extensive experiments, we show the efficacy and efficiency of \method on  exploratory graph analysis and graph classification tasks, where it significantly outperforms unsupervised baselines and achieves competitive performance with modern supervised GNNs, while achieving 
the best trade-off between accuracy and runtime.


\end{abstract}

\begin{IEEEkeywords}
attributed graphs, spectral embedding, graph filters, band-pass, density of states 
\end{IEEEkeywords}

\section{Introduction}
\label{sec:intro}

Graphs are widely used to model structured data from different domains such as chemistry \cite{wale2008comparison}, biology \cite{journals/bioinformatics/Przulj10}, cybersecurity \cite{duen2011polonium}, finance \cite{ribeiro2019shaping}, etc. 
The effectiveness and popularity of data-driven machine learning algorithms has necessitated  expressive vector representations of different kinds of complex data, 
and graphs are no exception.
Different from images or text, graphs pose novel challenges in finding effective representations as
 graph databases may contain graphs that vary in 
 size and structure, and do not necessarily exhibit alignment (i.e. correspondence) between the nodes of different graphs. 

Formally, we want to design a function $R: {G} \mapsto \z_G \in \R^D$, where $D$ is a fixed embedding size that does not depend on the input graph size.
Ideally, given a graph database with $N$ graphs (with $n$ nodes and $m$ edges per graph on average), we want $R$ to be 
($i$) \textit{permutation and size invariant}; where graphs with similar structure and label/attribute distribution have similar embeddings irrespective of node ordering and number of nodes,
($ii$) \textit{flexible}; that leverages information from node labels and/or multi-attributes as well as edge weights,
($iii$) \textit{multi-scale}/{\em glocal}; that can capture local/microscopic, mesoscopic, as well as global/macroscopic properties of a graph, and
($iv$) \textit{task-agnostic/unsupervised}; that can produce embeddings independent of any downstream task or related class labels, where not being tied to a specific task allows embeddings to be general-purpose for use e.g. in graph mining and exploratory data analysis. 
In addition, as with any algorithm, we want $R$ to be ($v$) \textit{efficient} and \textit{scalable} to large graphs (large $n$, $m$) as well as large databases (large $N$). Finally, $R$ that can produce one embedding at a time ($vi$) \textit{independently per graph} (as opposed to ``collective processing'') may be desirable, which allows on-the-fly embedding per incoming graph in streaming settings, 
as well as embarrassing parallelization for speed. 

\begin{figure}[!t]
		\vspace{-0.1in}
	\centering
	\begin{tabular}{cc}
			\hspace{-0.15in}\includegraphics[width=0.48\linewidth]{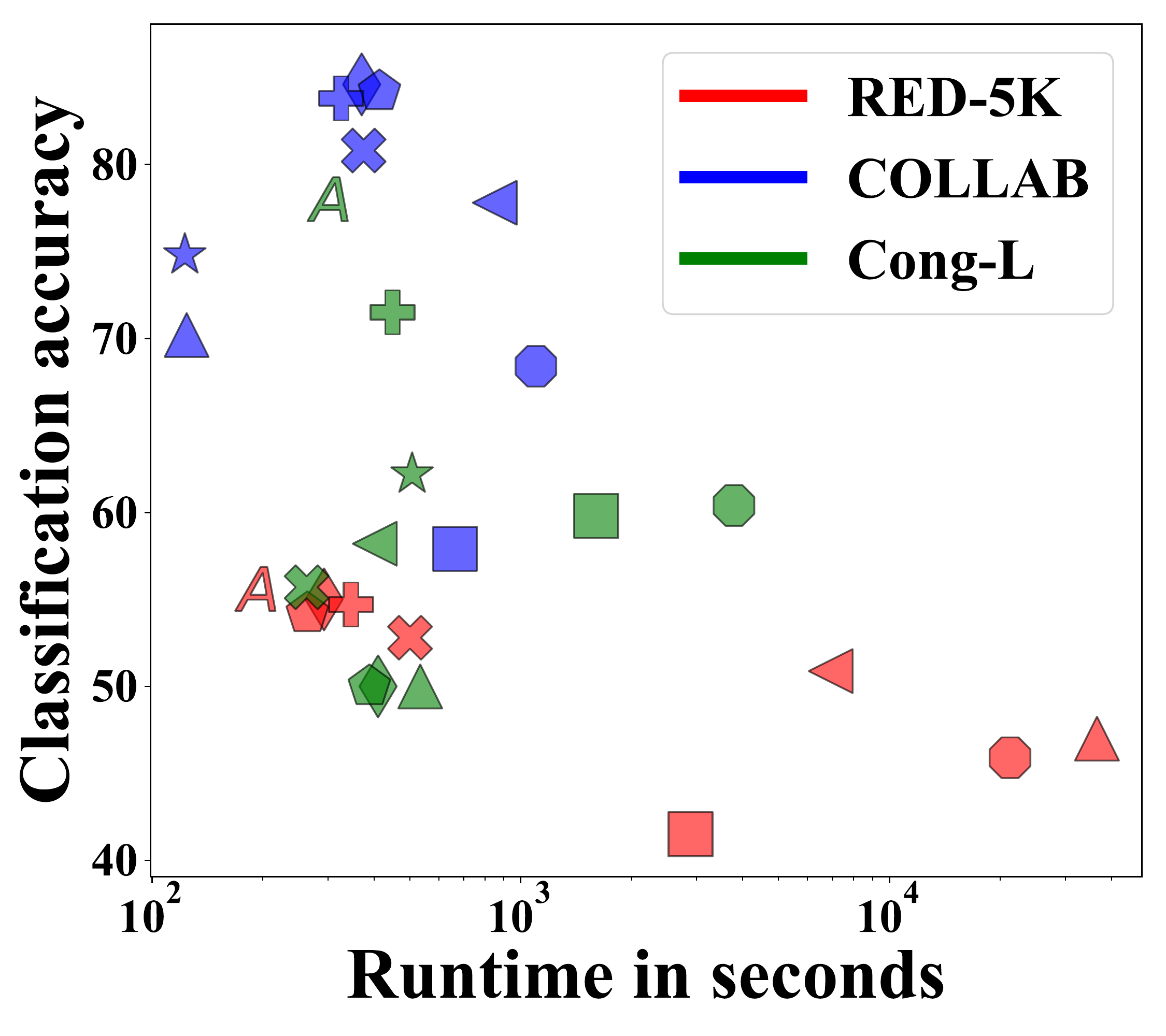}	& \hspace{-0.15in}\includegraphics[width=0.48\linewidth]{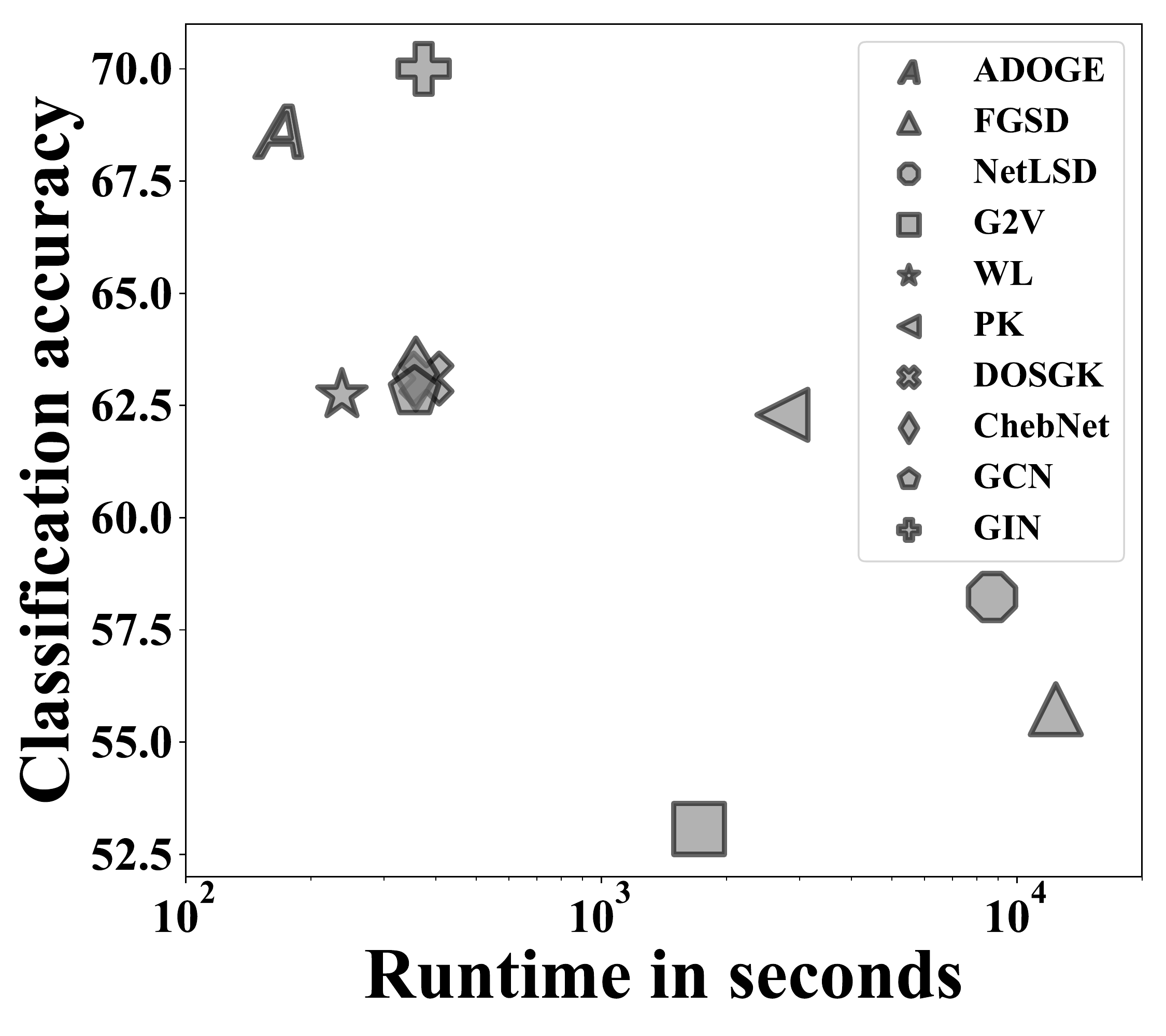} 
	\end{tabular}
	\vspace{-0.1in}
	\caption{\method achieves superior runtime-performance trade-off. (left) Runtime (log-scale) vs. accuracy on three individual datasets (with largest $N$, $\max(m)$ and the largest synthetic dataset). (right) Average runtime vs. graph classification accuracy across datasets.
		\label{fig:crown}}
	\vspace{-0.2in}
\end{figure}

Spectrally-designed embeddings are a popular class of techniques based on the graph eigenspectrum \cite{van2010graph},
as it captures key structural graph properties, such as cuts \cite{pothen1990partitioning},
random walk stationarity \cite{fill1991eigenvalue},
dynamical
processes and
epidemic thresholds \cite{chakrabarti2008epidemic}, 
diameter, connectedness, clustering, etc. \cite{jin2020spectral}.
However, the high complexity of computing the eigenspectrum exactly has proven to be a barrier for creating spectrum-based graph embeddings. Moreover, while the eigenspectrum can capture important topological properties, blending in node attributes/labels into spectrally-designed embeddings is nontrivial.

In this paper, we leverage {fast} algorithms for approximating the spectral density  of a graph \cite{dong2019network}, and use it to {independently}  construct {unsupervised} graph embeddings 
that are {permutation and size invariant}, {flexible} and {multi-scale}. Here, the focus is on representing the entire spectrum of the graph, which helps capture any arbitrary ``band'' of eigenvalues ({band-pass}), rather than only the extremal eigenpairs ({low/high pass}).

\vspace{-0.05in}
\subsection{Prior Work}
\label{ssec:prior}

Table~\ref{tab:salesman} gives a comparison with three categories of relevant prior work in the context of desired properties for a graph embedding. These existing work do not satisfy one or more of the aforementioned properties ($ii$)--($vi$) as we discuss next.

\subsubsection*{Unsupervised explicit graph embedding (UEGE)}

Several  unsupervised methods construct an explicit vector representation for each graph. Among those, {spectrum-based} methods have gained popularity in recent years.
FGSD \cite{verma2017hunt} treats a graph as a collection of spectral distances between its vertices.
NetLSD \cite{tsitsulin2018netlsd} represents a graph as a collection of heat traces of the graph at several time points.
Both methods are effective at capturing local and global structural properties of a graph, however, they ignore node labels and attributes.
graph2vec \cite{narayanan2017graph2vec} creates Weisfeiler-Lehman (WL) subtree based features, 
and learns an embedding of the graph trained to predict the existence of subtrees in the graph.
It admits node labels, but ignores node attributes as well as edge weights.



\begin{table}[!t]
	\vspace{-0.1in}
	\footnotesize
	\caption{ \textbf{\method satisfies all properties}, while prior work
		miss one or more of the input graph or embedding properties.\label{tab:salesman}}
	\begin{center}
		\vspace{-0.1in}
		{\renewcommand{\arraystretch}{1}%
			\begin{tabular}{l|l|p{0.1cm}p{0.1cm}p{0.0cm}|p{0.1cm}p{0.1cm}p{0.1cm}p{0.1cm}|}
		&	& \multicolumn{3}{c|}{\textbf{Inp. graph}} & \multicolumn{4}{c|}{\textbf{Embedding}} \\
			\hline
		&	\diagbox{\hspace{8mm}Method\hspace{-4mm}}{\raisebox{-1cm}{\rotatebox{90}{\vspace{5mm}Property}}}
			& \raisebox{-9mm}{\rotatebox{90}{\hspace{1mm}Node \textbf{labels}}}
			& \raisebox{-9mm}{\rotatebox{90}{\hspace{1mm}Node \textbf{attributes}\hspace{1mm}}} 
			& \raisebox{-9mm}{\rotatebox{90}{\hspace{1mm}Edge \textbf{weights}}} 
			& \raisebox{-9mm}{\rotatebox{90}{\hspace{1mm}\textbf{Band}-pass}}
			& \raisebox{-9mm}{\rotatebox{90}{\hspace{1mm}\textbf{Task-agnostic}}} 
			& \raisebox{-9mm}{\rotatebox{90}{\hspace{1mm}\textbf{Scalable}}}
			& \raisebox{-9mm}{\rotatebox{90}{\hspace{1mm}\textbf{Indpt.} per graph}\hspace{1mm}}  \\ 
			\hline
			
{\multirow{2}{*}{\rotatebox[origin=c]{90}{UEGE\hspace{-0.6mm}}}} 	&	\fgsd \cite{verma2017hunt}, 	\netlsd \cite{tsitsulin2018netlsd} &  &   &\textcolor{green}{\CheckmarkBold}  & & \textcolor{green}{\CheckmarkBold} & \textcolor{green}{\CheckmarkBold} & \textcolor{green}{\CheckmarkBold}\Tstrut \\ 						
			
	&		
			\gvec \cite{narayanan2017graph2vec} & \textcolor{green}{\CheckmarkBold} &    & & &\textcolor{green}{\CheckmarkBold} & \textcolor{green}{\CheckmarkBold}   &\Bstrut \\ 	
			
				\hline
{\multirow{4}{*}{\rotatebox[origin=c]{90}{GK}}}	&		\wl \cite{journals/jmlr/ShervashidzeSLMB11}, \wloa \cite{conf/nips/KriegeGW16}  & \textcolor{green}{\CheckmarkBold} &    & & & \textcolor{green}{\CheckmarkBold}&  &\Tstrut  \\ 	
	&		
			\sage \cite{sage}, \pk \cite{journals/ml/NeumannGBK16}  & \textcolor{green}{\CheckmarkBold} &  \textcolor{green}{\CheckmarkBold}  & \textcolor{green}{\CheckmarkBold} & &\textcolor{green}{\CheckmarkBold} &  &   \\ 	
			
	&		\retgk \cite{conf/nips/ZhangWXHN18}  &  &   & \textcolor{green}{\CheckmarkBold} & &\textcolor{green}{\CheckmarkBold} &  &   \\ 	
			
			
	&		\dos \cite{huang2021density} &  &    & \textcolor{green}{\CheckmarkBold}& \textcolor{green}{\CheckmarkBold} & \textcolor{green}{\CheckmarkBold}&   &\Bstrut \\
			
				\hline
		
{\multirow{2}{*}{\rotatebox[origin=c]{90}{GNN}}}	&		\gcn \cite{Kipf2016tc}, \gin \cite{Xu2019ty} & \textcolor{green}{\CheckmarkBold}  &  \textcolor{green}{\CheckmarkBold}   & \textcolor{green}{\CheckmarkBold} & & & \textcolor{green}{\CheckmarkBold}  &\Tstrut  \\
			
			
	&			\cheb \cite{conf/nips/DefferrardBV16}, \caley \cite{journals/tsp/LevieMBB19}  & \textcolor{green}{\CheckmarkBold} & \textcolor{green}{\CheckmarkBold}   & \textcolor{green}{\CheckmarkBold}& \textcolor{green}{\CheckmarkBold} & &  \textcolor{green}{\CheckmarkBold} &\Bstrut  \\
			
			\hline
	&		\method [this paper]  & \textcolor{green}{\CheckmarkBold} & \textcolor{green}{\CheckmarkBold}   & \textcolor{green}{\CheckmarkBold}& \textcolor{green}{\CheckmarkBold} & \textcolor{green}{\CheckmarkBold}&  \textcolor{green}{\CheckmarkBold} & \textcolor{green}{\CheckmarkBold}\Tstrut\Bstrut \\
			\hline
		\end{tabular} }
		
	\end{center}
	\vspace{-0.3in}
\end{table}

\subsubsection*{Graph kernels (GK)}
Due to the existence of many effective distance measures between graphs, {graph kernels} are a more widely studied method of graph representations \cite{journals/ans/KriegeJM20}. While most popular kernels
are effective at capturing characteristics of the graph structure, only a few, including the Propagation Kernel (PK) \cite{journals/ml/NeumannGBK16} are able to factor in edge weights, node labels and continuous node attributes (see Table 1 in \cite{journals/ans/KriegeJM20}).

Several graph kernels which use spectral properties have been developed in recent years.
{RetGK} \cite{conf/nips/ZhangWXHN18} represents each graph as a collection of node embeddings, where the node features are the return-probabilities of random walks of varying lengths.
{SAGE} \cite{sage} 
extends this idea to graphs with labeled and attributed nodes by appending each node embedding with its one-hot encoded label and/or attributes.
However, both these methods do not scale well for large graphs. Moreover, computing return probabilities of random walks tends to over-represent local features near a node, and often fails to capture global properties of the graph \cite{huang2021density}.
These issues are addressed by the 
Density Of States (DOS) GK, and its point-wise (i.e. node-level) extension (PDOS)\footnote{This is called LDOS in their paper, but we use PDOS to avoid confusion with our definition of LDOS in this paper.}, which uses Chebyshev polynomials to efficiently capture global properties of random walks, and uses fast approximation techniques \cite{dong2019network}.
However, despite their efficiency, they are limited to plain graphs and do not admit node labels or attributes.

Moreover, although graph kernels have proven effective at modelling graph structure, and in some cases node labels and attributes, for many kernel methods, computing an $N$$\times$$N$-sized kernel matrix can be restrictive in terms of both time and space, which do not scale to large databases with many graphs.

\subsubsection*{Graph Neural Networks (GNNs)}

While most existing unsupervised embedding and kernel methods are ill-equipped to handle continuous node attributes, GNNs are able to leverage such data to a great extent.
However, deep parameterized models come with their own drawbacks. They are resource-hungry, not task-agnostic, and can be slow to train.
Moreover, when viewed through a spectral lens \cite{balcilar2021analyzing}, most neighborhood-aggregation based GNNs such as {GCN} \cite{Kipf2016tc} and {GIN} \cite{XuHLJ19} can only act as low-pass or high-pass filters on a graph spectrum. Only spectrally designed GNNs such as {ChebNet} \cite{conf/nips/DefferrardBV16} and {CaleyNet} 
\cite{journals/tsp/LevieMBB19} can act as band-pass filters.


A perhaps subtle characteristic of graph embedding methods is \textit{independent} versus dependent/collective processing of the graphs. By design, all GNN-based methods including graph2vec require collective processing due to end-to-end training.
WL and PK respectively obtain the compressed labels and histogram bins based on all graphs which makes them dependent. RetGK, DOSGK, and SAGE obtain graph-level embeddings through kernelizing the set of node-level embeddings, which is of different sizes across graphs, and hence they are inherently bound to create $N$$\times$$N$ pairwise kernel values rather than an explicit/independent embedding for each graph.

\vspace{-0.05in}
\subsection{Our Contributions}
We propose \method (Attributed DOS-based Graph Embedding), for extremely fast unsupervised embedding of attributed graphs that is permutation and size invariant, flexible, and multi-scale, which is produced independently per graph.

Our main technical contributions are as follows:

\noindent\sbt~ {\bf New graph-level embedding algorithm:}
We introduce a new  spectrally-designed graph embedding approach, called \method,
that leverages the whole (eigen)spectrum of a graph. 
\method capitalizes on recent algorithms that can efficiently approximate the (local) density of states (L)DOS \cite{dong2019network}, extending to attributed graphs for the first time.

\noindent\sbt~ {\bf Desired characteristics:} 
Thanks to efficient approximations, \method is extremely fast.
It can handle node labels, continuous multi-attributes, and edge weights.
Leveraging the whole spectrum, it enables variable band-pass filtering as well as features that capture multi-scale properties.
Further, it processes each graph independently of others, which makes it amenable for streaming scenarios as well as parallelization.

\noindent\sbt~ {\bf Exploratory graph analysis: } 
\method is not tied to any specific objective, which makes it suitable for both un/supervised tasks.
In fact, our embedding features lend themselves to various interpretations, related to graph signal convolution,
random walks, and band-filters, which
%
%
prove useful in data mining and exploratory analysis of real-world graph datasets as we show through experiments.

\noindent\sbt~ {\bf Efficacy and Efficiency:~} 
Extensive experiments show that \method is
on par with or superior to all unsupervised baselines,
 and competitive against modern supervised GNNs on graph classification tasks.
Notably, 
it achieves 
the best runtime-accuracy trade-off.
(See Fig. \ref{fig:crown}.)

\hide{
\bit
\item {\bf New graph-level embedding algorithm:}
introduce a new spectrally-designed 
density of states (DOS), a.k.a. spectral density, 

whole spectrum 

capitalizing on new algorithms \cite{dong2019network}

\item  {\bf Desired characteristics:} 
- fast 
- can handle labels, continuous multi-attributes, edge weights
	how attributes distribute over / align with the graph structure,
	rather than a collection of node attributes
- whole spectrum,--> allows for variable band-pass filters, as opposed to low/high-pass only

\item  {\bf Exploratory graph analysis: } 
interpretations; relation to graph signal convolution,
random walks, band-filters, etc.
since embedding not tied to a specific objective, ameneble for unsupervised tasks such as exploratory graph analysis...

 interpretable/explainable features which prove useful in data mining and exploratory analysis of graph datasets. ..

\item {\bf Efficacy and Efficiency:~} 
on par with all unsupervised baselines, and competitive against supervised modern GNNs on graph classificaiton tasks.
Super-fast, less than 1 sec 
linearly scalable w.r.t. graph size
\eit
}

\textbf{Reproducibility}: We share all datasets and source code at \url{https://github.com/sawlani/A-DOGE}.











\section{Problem Statement \&  Preliminaries }
\label{sec:prelim}


\noindent \textbf{Notation.~}
We denote scalars, vectors, matrices and sets by lowercase ($x$), lowercase boldface ($\x$), uppercase boldface ($\X$), and calligraphic ($\mX$) letters, respectively.
$\X_{:j}$ 
and $\X_{ij}$ refer to the $j$-th column 
and the $(i,j)$-th entry of a matrix.

We consider undirected, weighted node-attributed graphs $G=(\mV, \mE, \X, \mathcal{A})$ where $\mV=\{v_1,\ldots,v_n\}$ denotes the set of $n$ nodes, and $\mE \subseteq \mV \times \mV$ denotes the set of $m$ edges. 
$\W$ depicts the weighted adjacency matrix where $\W_{ij} > 0$ if $(v_i,v_j) \in \mE$, and $0$ otherwise. 
$\X$ is the $n \times d$ node-attribute matrix, where 
$\mathcal{A}=\{a_1,\ldots,a_d\}$ denotes the set of $d$ attributes, with $dom(a_j)$ depicting the domain of attribute $a_j$.
In terms of Graph Signal Processing (GSP) terminology, any $\x=\X_{:j}$ can be thought as a \textit{graph signal} on the nodes, with one scalar per node.

\begin{problem}[Unsupervised Graph-level Embedding]
	\textit{\em \bf Given} a set of undirected, weighted and node-attributed/labeled graphs 
	$\mathcal{G} = \{G_1, \ldots, G_N\}$, for 
	 $G_i=(\mV_i, \mE_i, \X_i, \mathcal{A})$, where
	 \bit
	 \item[($i$)] graphs in $\mathcal{G}$ can be of varying sizes,
	($ii$) there exists no particular correspondence between the nodes of different graphs, and
	($iii$) the (categorical and/or continuous) attributes and their domain are shared among all graphs,
	\eit
	  \textit{\em \bf Find} $D$-dimensional graph-level embedding $\z_G \in \R^D$ for each $G\in \mathcal{G}$ that captures both structural and attribute information.
\end{problem}

Let $\tW = \D^{-1/2} \W \D^{-1/2}$ denote the symmetrically normalized adjacency matrix, where 
$\D$ is the diagonal degree matrix with $\D_{i,i} = \sum_j \W_{i,j}$. Let
$\tL = I - \tW$ denote the Laplacian matrix, and $\bP=\D^{-1} \W$ the random walk matrix.
For a connected graph, $\tW$ has eigenvalues $-1=\lambda_0 < \lambda_1 \leq \ldots \leq \lambda_{n-1}=1$ with corresponding eigenvectors $\{\bu_k\}_{k=0}^{n-1}$. 
$\tW$ has the same set of eigenvectors as $\tL$ whose eigenvalues are the shifted set 
$\{\mu_k = 1- \lambda_k\}_{k=0}^{n-1} \in [0,2]$. 
 $\tW$ also shares the same eigenvalues as $\bP$.
As such, the spectral density function  has bounded support for these graph matrices.
 Following GSP convention, we
 refer to the eigenvalues as the {\em graph frequencies}.
 

In this work we use $\tW$ 
as the so-called {graph shift operator} $\bS$ which generalizes to any symmetric matrix of a graph.
Let $\bS = \U \Lam \U^T$ depict the eigendecomposition, where $\Lam := \text{diag}([\lambda_1 \ldots \lambda_n])$ and 
$\U = [\bu_1 \ldots \bu_n]$.

\begin{definition}[Graph spectrum]
The \textbf{spectrum} of a graph is composed of	the set of graph eigenvalues together with their multiplicities of the (normalized) adjacency matrix.
\end{definition}

\noindent  \textbf{Graph Fourier transform.~}
 The graph Fourier transform (GFT) of a graph signal $\x \in \R^n$ 
is defined as the projection 
\vspace{-0.05in}
\begin{align*}
\hy = \mathcal{F}(\x) = \U^T \x 
\vspace{-0.05in}
\end{align*}
and the inverse GFT of $\hy \in \R^n$ is given as 
\vspace{-0.05in}
\begin{align*}
\x = \mathcal{F}^{-1}(\hy) =  \U \hy
\vspace{-0.05in}
\end{align*}
\noindent  \textbf{Graph filtering.} A {graph filter} is an operation on a graph signal with output in the graph frequency domain, that is,
\vspace{-0.05in}
\beq
\label{eq:gflt}
\hy_{{flt}} = \phi(\Lam) \hy \;,
\vspace{-0.05in}
\eeq
where $\phi(\Lam)$ is a diagonal matrix with filter {frequency response} values as its diagonal elements. 

\begin{definition}[Frequency Response Function (\frf)]
The \textbf{frequency response function} of a graph filter is written as
\vspace{-0.05in}
\beq
\label{eq:frf}
\phi: \mathbb{C} \mapsto \R , \;\; \lambda_i \rightarrow \phi(\lambda_i) \;,
\vspace{-0.05in}
\eeq
which, simply put, assigns a scalar value $\phi(\lambda_i)$ to each graph frequency (i.e., eigenvalue) $\lambda_i$.
\end{definition}

By applying the inverse GFT on both sides of Eq. \eqref{eq:gflt}, we can get the filter output in the node domain as
\vspace{-0.05in}
\begin{align*}
\x_{{flt}} = \U \phi(\Lam) \hy = \U \phi(\Lam) \U^T \x = \phi(\bS) \x \;.
\vspace{-0.05in}
\end{align*}

\noindent  \textbf{Signal convolution.~}
Graph convolution of two signals, say $\x$ and $\x'$, each in $\R^n$, yields another signal $\bc \in \R^n$ as 
\vspace{-0.05in}
\begin{align*}
\bc_{\x,\x'}  = \x *_G \x' = \U ( \U^T \x \odot \U^T \x' ) = \sum_{i=1}^n   \bu_i   (\bu_i^T \x) (\bu_i^T \x')
\vspace{-0.05in}
\end{align*}
where $\odot$ depicts the Hadamard product.
We can write the Fourier transform of the convolution as 
\vspace{-0.05in}
\beq
\label{eq:conv}
\mathcal{F}(\bc_{\x,\x'}) = \widehat{\bc}_{\x,\x'} = \{ (\bu_i^T\x)  (\bu_i^T\x') \}_{i=1}^n \;.
\vspace{-0.05in}
\eeq

\noindent  \textbf{Density of States.~}
{Spectral density} is the overall distribution of the eigenvalues as induced by any symmetric $n \times n$ graph matrix $\bS=\U \Lam \U^T$. It is also referred to as the {density of states (DOS)} in the physics literature, reflecting the number of states at different energy levels \cite{wang2001efficient}. Formally,

\begin{definition}[Density of States (DOS)] DOS or the spectral density induced by $\bS$ is the density function
\vspace{-0.05in}
\beq
\label{eq:dos}
f(\lambda) = \frac{1}{n}\sum_{i=1}^n \delta(\lambda - \lambda_i)\;,
\vspace{-0.05in}
\eeq
where $\delta(\cdot)$ is the Dirac delta function.
\end{definition}


\begin{definition}[Local Density of States (LDOS)] Likewise, for \textit{any} input vector $\bv \in \R^n$, LDOS is given as
\vspace{-0.05in}
	\beq
	\label{eq:ldos}
	f(\lambda; \bv) = \sum_{i=1}^n |\bv^T \bu_i|^2 \delta(\lambda - \lambda_i)\;.
\vspace{-0.05in}
	\eeq
\end{definition}


The following related equalities can be derived easily respectively for DOS and LDOS.
\vspace{-0.05in}
\beq
\label{eq:dosint}
\int \phi(\lambda) f(\lambda) = \sum_{i=1}^n \phi(\lambda_i) = \text{trace}(\phi(\bS))
\vspace{-0.05in}
\eeq
\vspace{-0.05in}
\beq
\label{eq:ldosint}
\int \phi(\lambda) f(\lambda; \bv) = \sum_{i=1}^n \phi(\lambda_i) (\bv^T \bu_i) (\bu_i^T \bv) =  \bv^T \phi(\bS) \bv
\vspace{-0.05in}
\eeq

{\bf Scaling (L)DOS.~} The extremal (i.e., a few top largest or smallest) eigenpairs of various graph matrices have been associated with important graph characteristics, such as small-cut partitions \cite{pothen1990partitioning}, 
convergence rate of random walks to stationarity \cite{fill1991eigenvalue},
unfolding of dynamical
processes and
epidemic thresholds \cite{chakrabarti2008epidemic}, 
 etc.
Obtaining those few eigenpairs is also computationally easy.
On the other hand, (L)DOS provides the distribution of the \textit{entire} spectrum, which opens the door for the analysis of graph properties that are not evident from only the extremal eigenpairs.
However, 
computing {all} $n$ eigenvalues and eigenvectors of a graph with $n$ nodes is considerably more demanding.
Therefore, analyzing large graphs through their density of states has been obstructed by the lack of scalable algorithms, until recently. 

In their award-winning work, 
Dong {\em et al.} \cite{dong2019network} introduced highly-efficient approximation algorithms to compute spectral densities, scalable to graphs with as large as tens of millions of nodes and billions of edges.
Their main focus has been scaling the computation of these functions, with approximation-error analysis on plain graphs. 
In this paper,
we capitalize on their work for speed and extend it to leverage node attributes for the first time for fast, attributed graph-level embedding. 

\section{Graph-level Embedding with \method}
\label{sec:method}

\subsection{Motivation} 

Our spectrally-designed \method derives graph-level features based on the node attributes and the {\em entire} spectrum of $\tW$ 
(can be other symmetric graph matrix, w.l.o.g. referred as $\bS$, see \S\ref{sec:prelim}), 
where the spectrum is composed of \textit{all} the eigenvalues. 
Before delving into details, we discuss the motive for using the full spectrum and present an illustrative example.

\textbf{Why the entire spectrum?~} We design graph-level features based on {all} of the eigenpairs of a graph matrix for two primary reasons.
First,  a large number of studies have found that the full eigenvalue spectra of different classes of real-world networks differ considerably \cite{jin2020spectral,farkas2001spectra,banerjee2008spectral,mcgraw2008laplacian}. This suggests that the spectra can play a key discriminative role.
Second, real-world networks are observed to exhibit localization on low-order eigenvectors, which are those eigenvectors associated with the non-extremal eigenvalues (in the sense of being the largest or smallest), but that are ``buried'' further down in the eigenvalue spectrum \cite{cucuringu2011localization}.
Notably, they capture mesoscopic inhomogeneity in networks which is defined  as topologically distinct groupings of nodes, from few nodes to large modules, communities, or different interconnected subnetworks
\cite{mitrovic2009spectral}.


%

{\bf Illustrative example:~}
To illustrate the valuable information that non-extremal eigenpairs carry,
we present a visual analysis of low-order eigenvector localization using the \mig graph (See \S\ref{ssec:setup}). It consists of the counties across 49 mainland U.S. states as nodes, and an  edge depicts the total number of people that migrated between two counties during 1995-2000 
\cite{cucuringu2011localization}.

\begin{figure}[!t]
\vspace{-0.1in}
	\centering
	\begin{tabular}{ccc}
\hspace{-0.2in}	\includegraphics[width=0.33\linewidth]{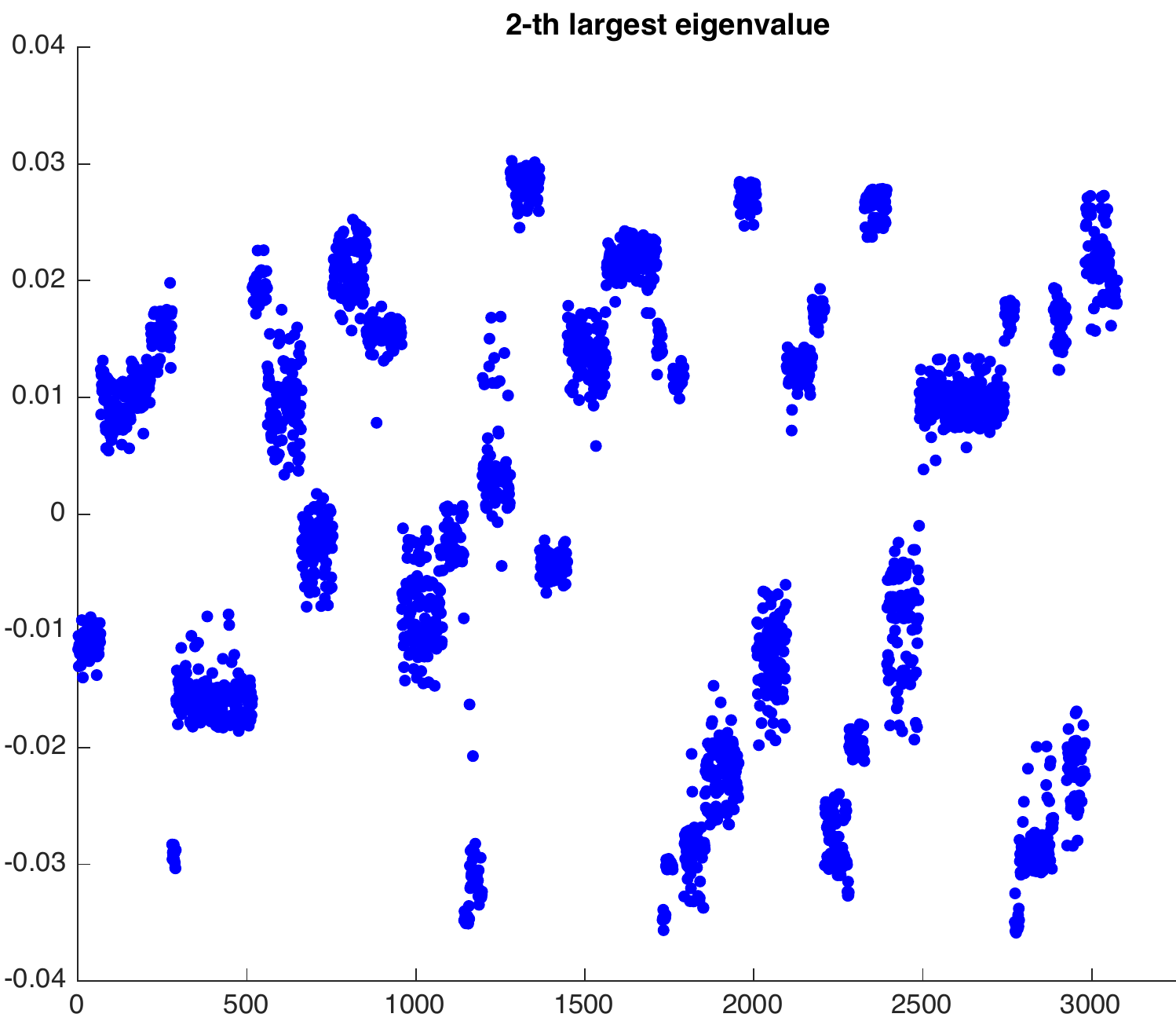} &
\hspace{-0.15in}	\includegraphics[width=0.33\linewidth]{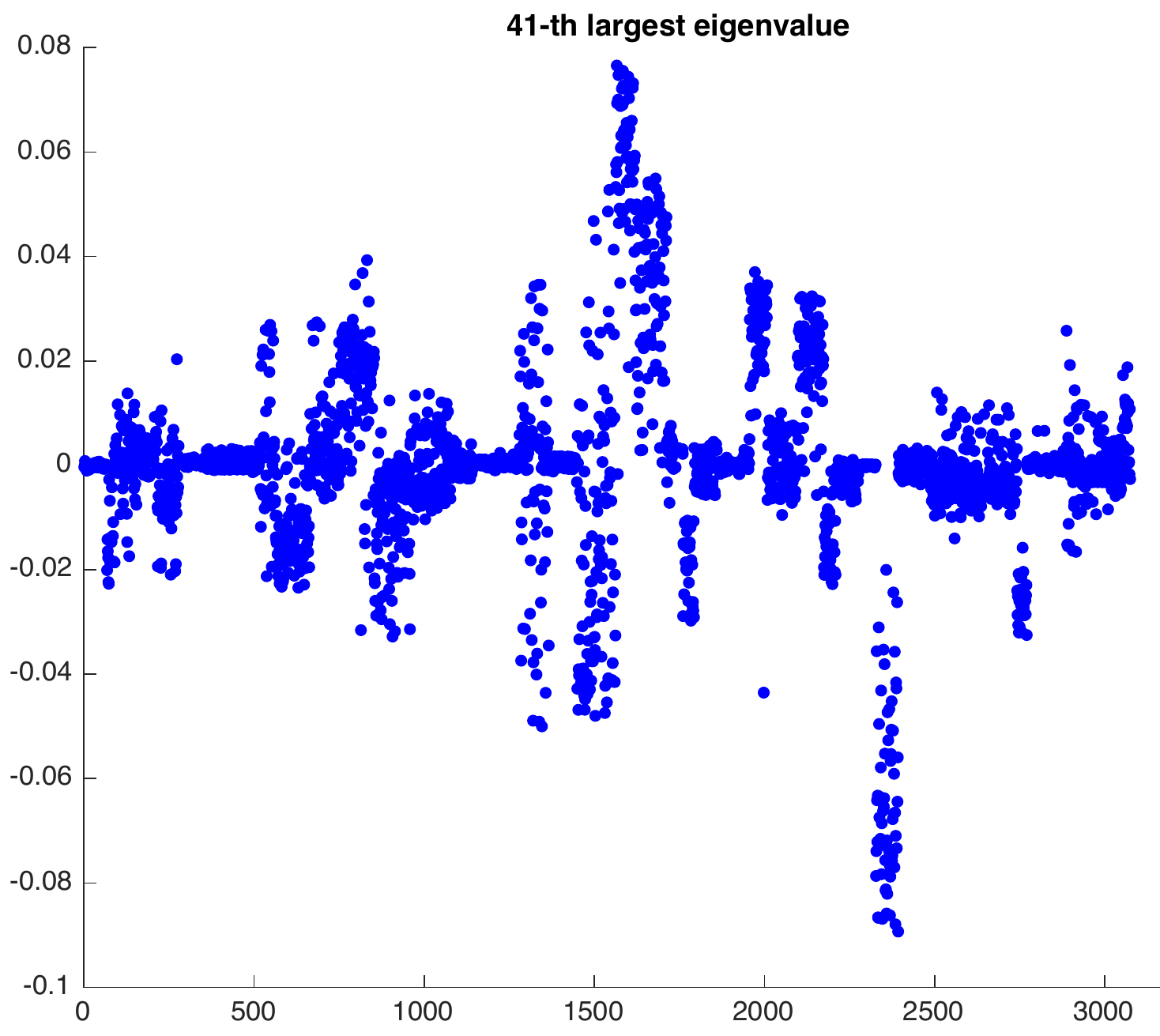} &
\hspace{-0.15in}	\includegraphics[width=0.33\linewidth]{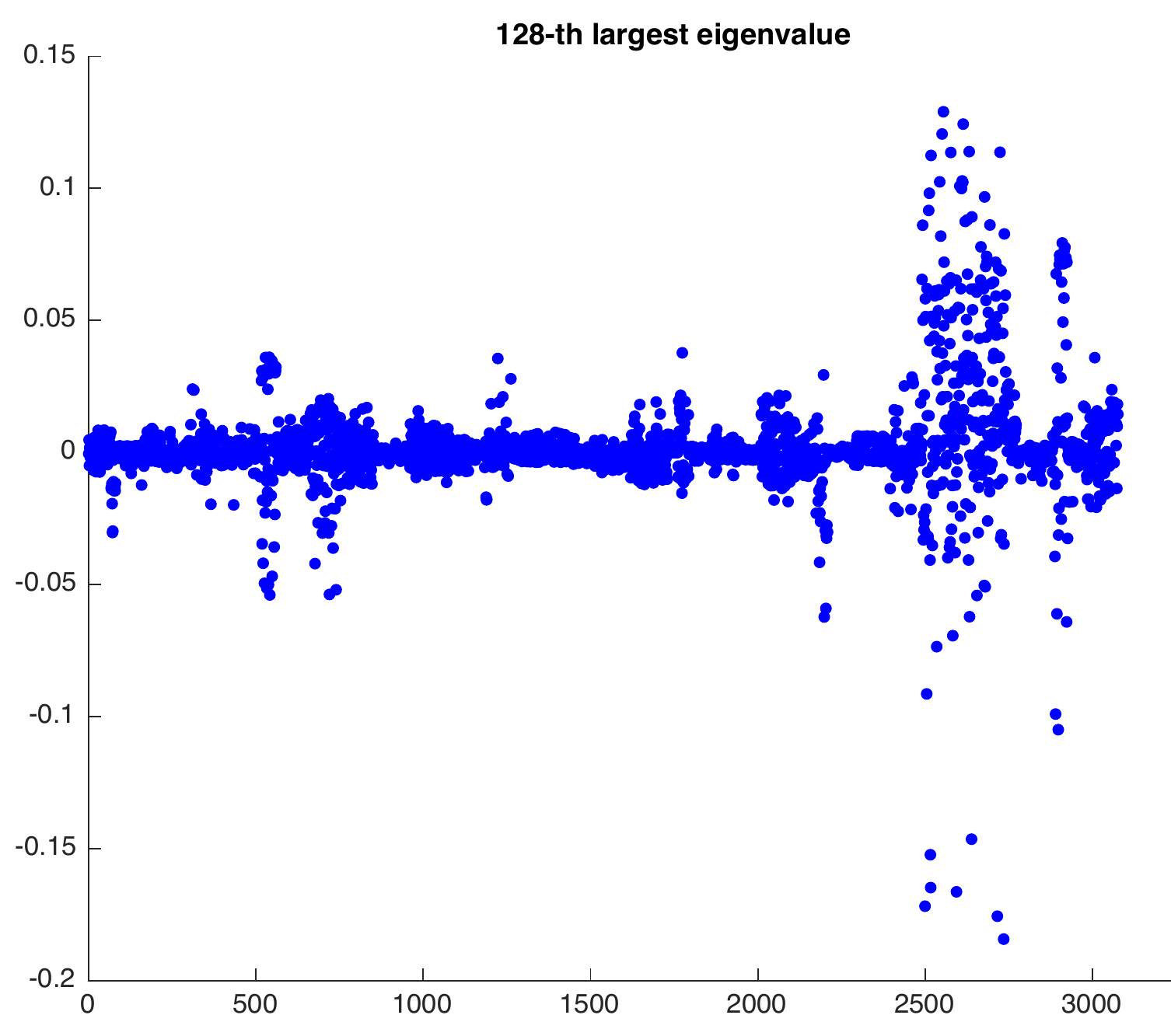} \\
\hspace{-0.2in}		\includegraphics[width=0.33\linewidth]{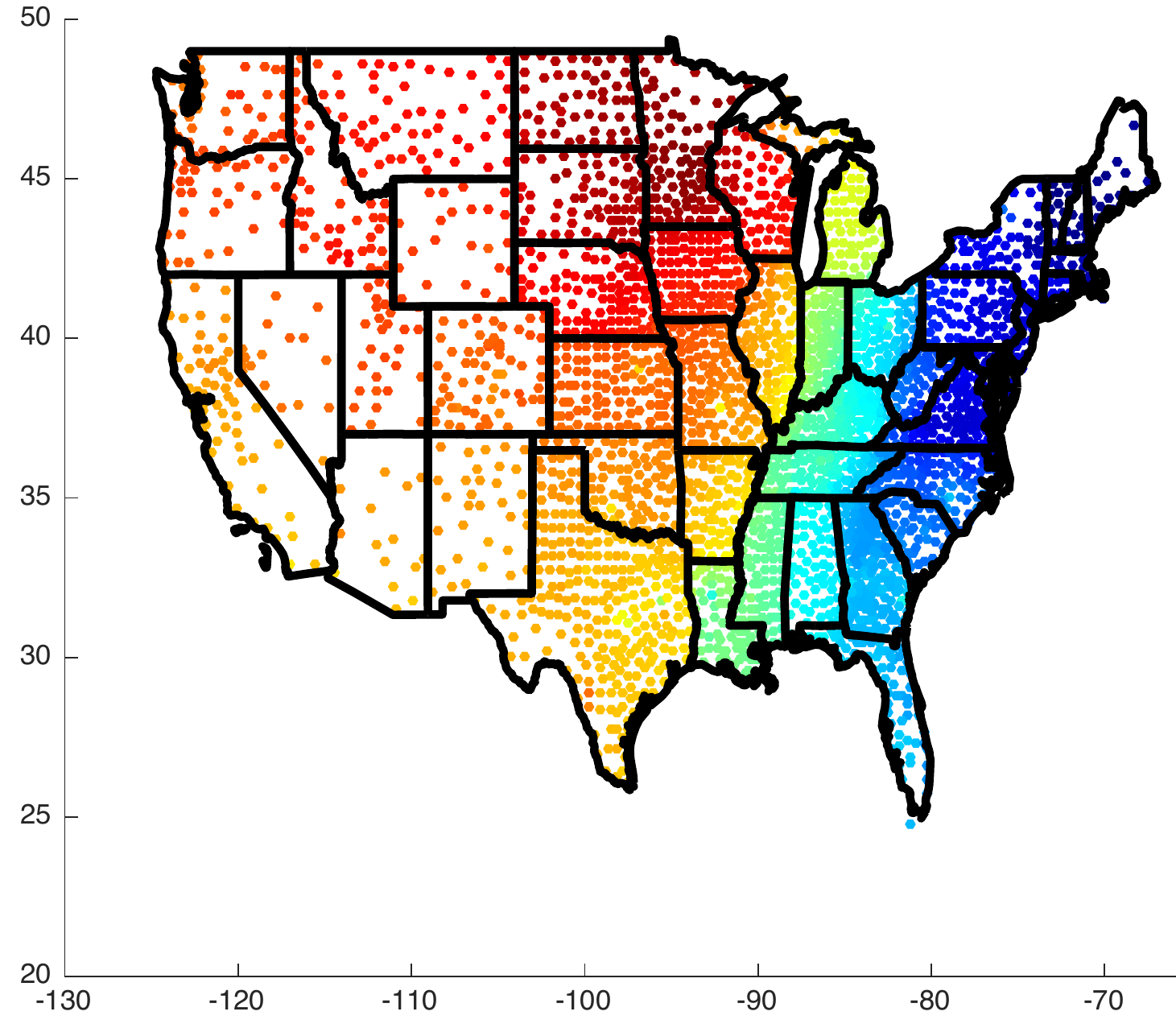}&
\hspace{-0.15in}	\includegraphics[width=0.33\linewidth]{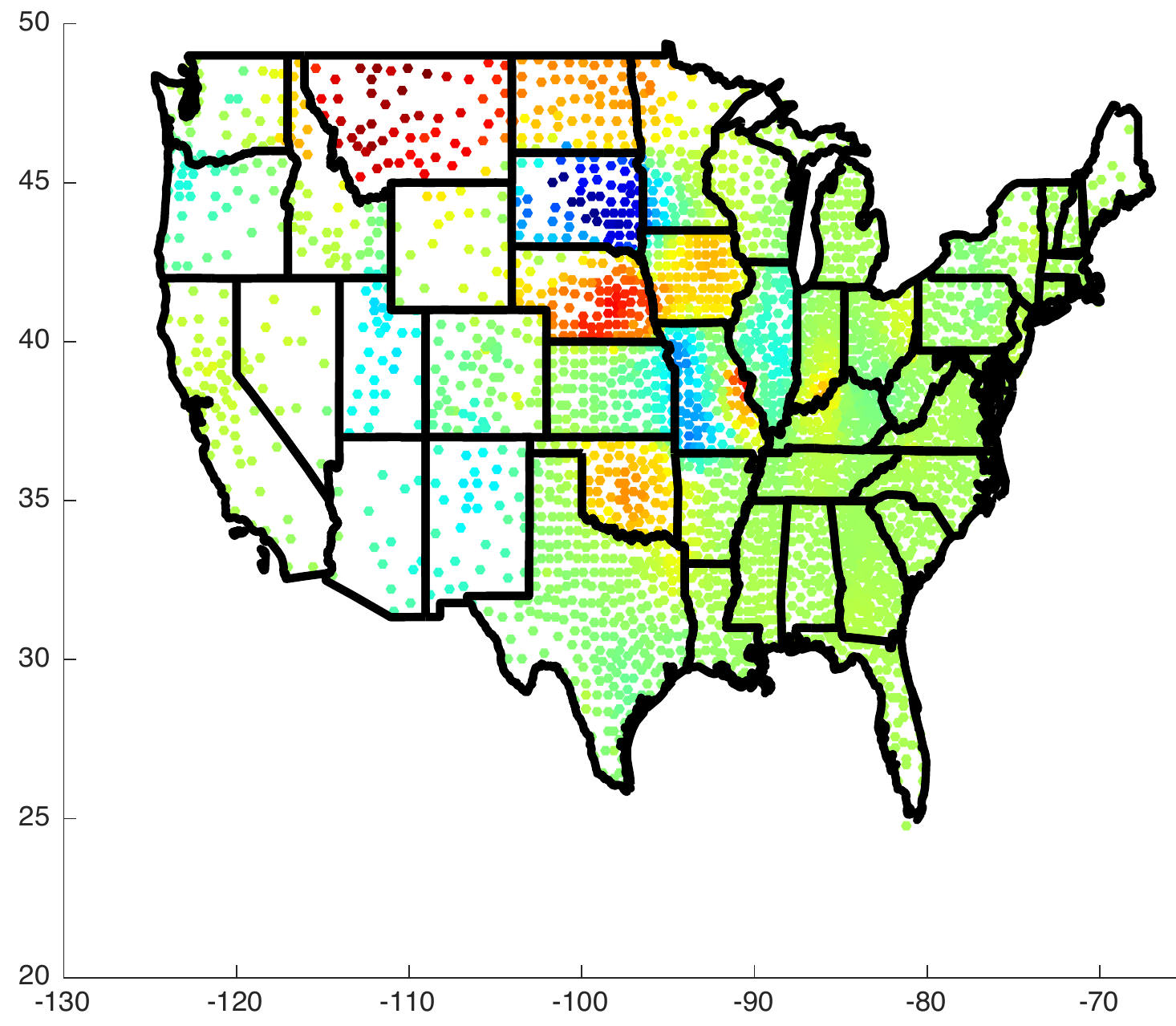}&
\hspace{-0.15in} \includegraphics[width=0.33\linewidth]{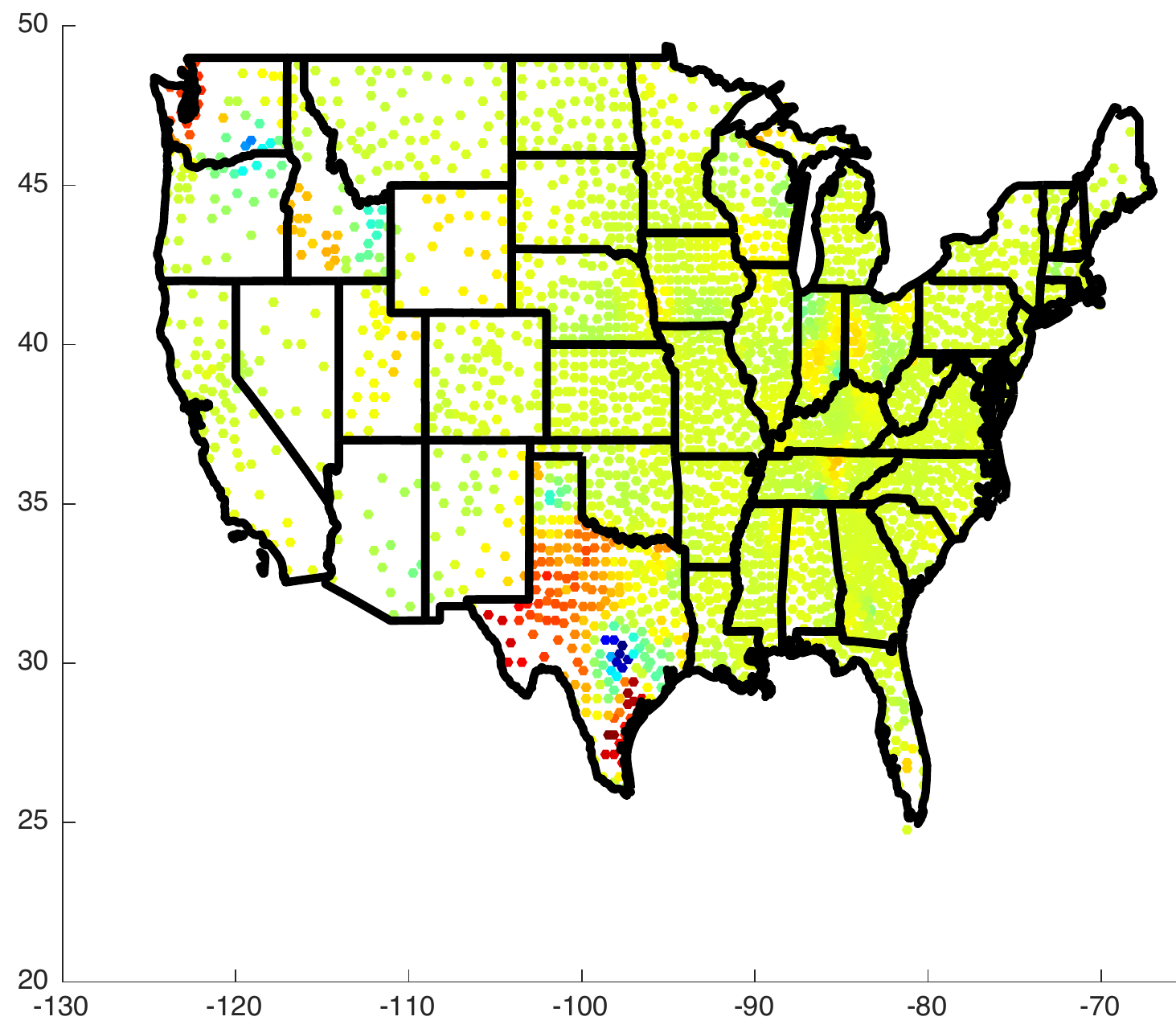}\\	
(a) 2nd, \textit{macro} & (b) 41st, \textit{meso} & (c) 128th, \textit{micro} 
	\end{tabular}
	\vspace{-0.1in}
	\caption{(top) Eigenvector entries (y-axis) versus node (i.e. U.S. county) index (x-axis) for the top (from left to right) 2nd, 41st, and 128th eigenvectors of the \mig graph (See \S\ref{ssec:setup}); (bottom) Eigenvector entries as heatmap (red:high to blue:low) for nodes (U.S. counties) shown in 2-d coordinates, solid black lines depict official U.S. state borders (best in color). \label{fig:maps}}
	\vspace{-0.2in}
\end{figure}

Eigenvector localization arises when most of the entries of an eigenvector are zero or near-zero, and 
implies that the nonzero components of the eigenvector coincide with a particular set of geometrically distinguished nodes in the graph.
Extremal eigenvectors typically exhibit low localization;
as shown in Fig. \ref{fig:maps}(a), the 2nd eigenvector 
has many non-zeros and mainly captures macroscopic properties, in this case, the graph cut depicting relatively fewer migrations between west- and east-coasts.
A lower-order eigenvector, namely the 41st in (b) reflects mesoscopic structure in terms of migration patterns in and around South Dakota.
128th eigenvector  
in (c) has even larger localization,
narrowing in a few counties within Texas near Austin, reflecting microscopic patterns.  
It is remarkable that the low-order eigenvectors align with geographical and political boundaries, carrying useful information at multiple scales.

%


 

\vspace{-0.05in}
\subsection{Spectrum as Histogram}

Density of states (DOS) or spectral density as given in Eq. \eqref{eq:dos} is a continuous probability density function
$f(\lambda)$ of the eigenvalues. We represent it with a histogram density estimator, denoted $h^{DOS}(\lambda)$ that partitions the eigenvalue range $[-1,1]$ for $\tW$ 
into 
 $B=2/\Delta$ disjoint bins of equal width $\Delta$.
Let us denote their centers by $\widetilde{\lambda}_b$ for $b \in \{1, \ldots, B\}$.
For any $i \in \{1,\ldots, n\}$, let $\text{Bin}(\lambda_i)$ denote the bin that $\lambda_i$ belongs to.
We define our DOS histogram features for a graph as follows.

\begin{definition}[DOS histogram features]
	DOS histogram is a $B$-dimensional vector, denoted $\hdos \in \R^B$, where 
	\vspace{-0.05in}
	\begin{align}
		\label{eq:dosfeat}
	\hdos(\widetilde{\lambda}_b) = \frac{1}{\Delta} \frac{\sum_{i=1}^n \I(\lambda_i \in \text{\em Bin}(\widetilde{\lambda}_b))}{n} \;, b \in \{1, \ldots, B\}
	\vspace{-0.05in}
	\end{align}
\end{definition}

\begin{figure}[!t]
	\vspace{-0.1in}
	\centering
	\begin{tabular}{c}
		\hspace{-0.25in}	\includegraphics[width=0.9\linewidth,height=1.1in]{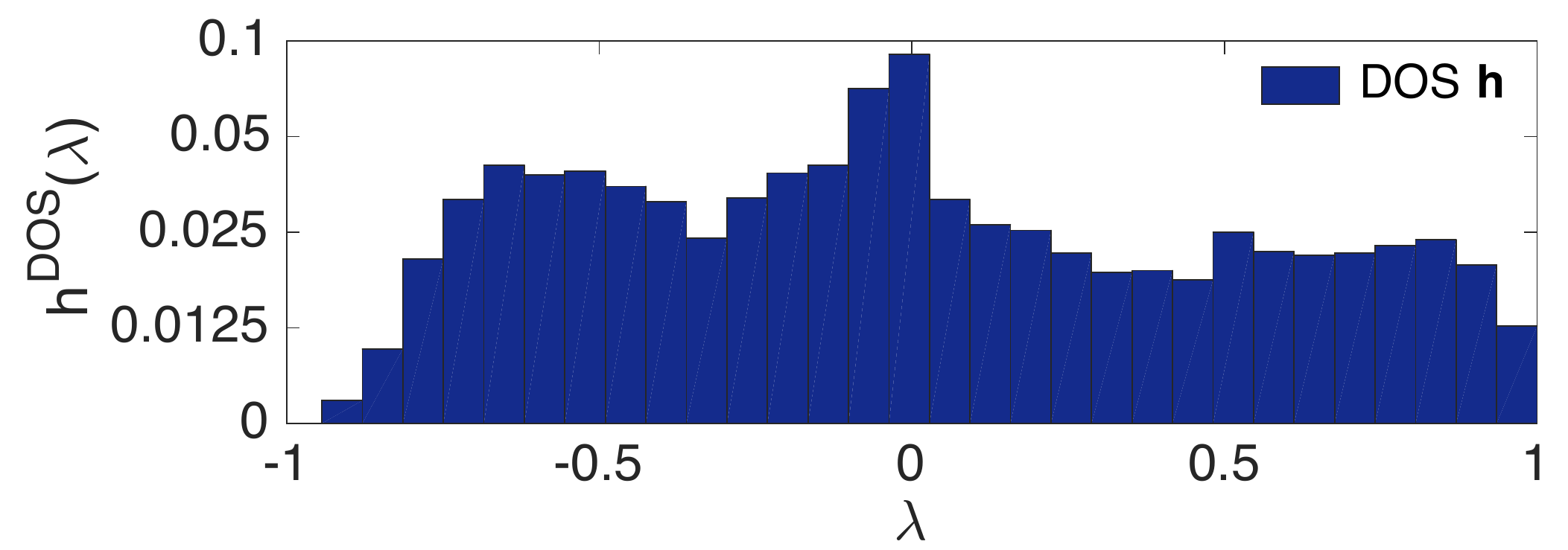} \\
		\hspace{0.15in}	
		\includegraphics[width=0.9\linewidth,height=1.15in]{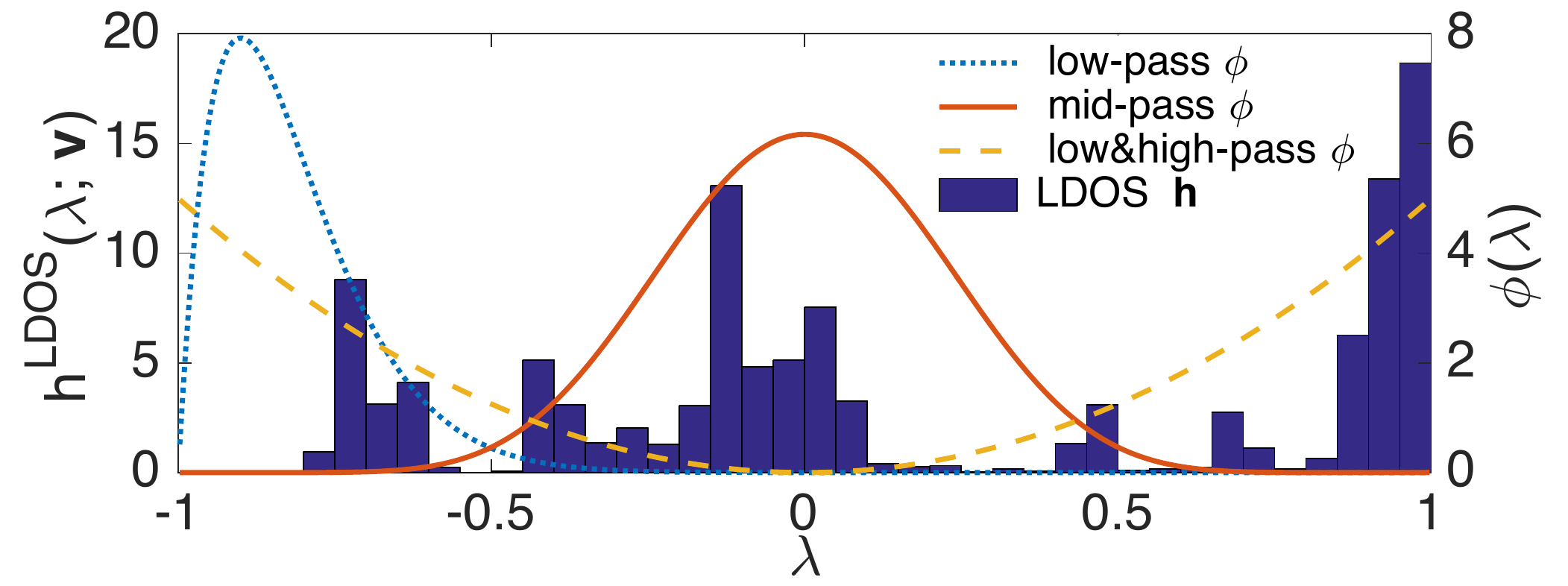} 
	\end{tabular}
	\vspace{-0.2in}
	\caption{Example (top) histogram density estimator of the spectral density, a.k.a. density of states (DOS), of a graph for symmetrically-normalized $\tW$ with eigenvalues (i.e. frequencies) in $[-1,1]$, (bottom) local density of states (LDOS) histogram for a given vector $\bv$. Also shown (in color) are three different frequency response functions depicted by curves over the spectrum. \label{fig:ldos}} 
	\vspace{-0.2in}
\end{figure}

We also represent the local density of states (LDOS) in Eq. \eqref{eq:ldos} similarly and define LDOS histogram features.
\begin{definition}[LDOS histogram features]
	For a given vector $\bv \in \R^n$, the LDOS histogram is a $B$-dimensional vector, denoted $\hldos_{\bv} \in \R^B$, where 
	\vspace{-0.05in}
	\beq
	\label{eq:ldosfeat}
	\hldos_{\bv}(\widetilde{\lambda}_b) = \frac{1}{\Delta} \frac{\sum_{i=1}^n |\bv^T \bu_i|^2\; \I(\lambda_i \in \text{\em Bin}(\widetilde{\lambda}_b))}{n} \;, \forall b 
	\vspace{-0.05in}
	\eeq
\end{definition}
Note that by abusing convention slightly, we use the word histogram to refer to Eq. \eqref{eq:ldosfeat} although it is not a normalized density mass function.
Fig. \ref{fig:ldos} shows examples to DOS (top) and LDOS (bottom) histograms with $B=40$ each.

Computing both histograms requires all of the eigenvalues $\lambda_i, i=\{1\ldots n\}$
for a graph with $n$ nodes. Further, LDOS requires all the corresponding eigenvectors $\bu_i$'s.
For even moderate size graphs, computing the complete set of eigenpairs is prohibitive.
Most recently, Dong {\em et al.} \cite{dong2019network} introduced fast and scalable approximation algorithms to estimate these spectral densities. Our work is inspired by and builds on their work to efficiently obtain both $\hdos$ and $\hldos$ based on the Gauss Quadrature and Lanczos (GQL) algorithm \cite{books/ox/07/GolubM07}.

On the other hand, both in \cite{dong2019network} and their follow-up work \cite{huang2021density},
$\bv = \be_i$ is used in Eq. \eqref{eq:ldos} to capture the spectral information about each particular node $i=\{1,\ldots,n\}$, called point-wise density of states (PDOS), where $\be_i$ is the $i$-th standard basis vector with $i$-th entry equal to 1 and 0 elsewhere. 
As such, both work are limited to plain graphs without node labels/attributes.
We extend the use of LDOS to attributed graphs for the first time, \textit{by setting} $\bv \in \R^n$ in Eq. 
\eqref{eq:ldosfeat} to \textit{capture a graph signal on the nodes associated with an attribute}.

Specifically, given a categorical or binary attribute $a_j$, we create a separate $\bv$ for each unique value $val \in dom(a_j)$ where $\bv_i := 1$ if $\X_{ij} = val$ and 0 otherwise.
For numerical attributes, we set $\bv := \overline{\X}_{:j}$ where $\overline{\X}$ denotes the column-wise standardized attribute matrix.
Notably, LDOS can be extended to 
 structural node-level attributes, such as degree or other node centrality measures, eccentricity, etc.



{\bf Interpreting LDOS.} There is an intuitive interpretation of a LDOS feature in Eq. \eqref{eq:ldosfeat}.
The term $\bv^T \bu_i$, that is the dot product between an attribute vector and a graph eigenvector, is to reflect the
 \textit{alignment} between attribute values and the structurally distinct group of nodes that the eigenvector captures.
 The better the alignment, the larger is the LDOS feature value for the bin that the corresponding eigenvalue falls into.

In addition to original LDOS, we also create {\em interaction} features between pairs of node attributes.
Accordingly, the coupled-LDOS histogram features are defined as follows.


\begin{definition}[cLDOS histogram features]
	For two input vectors $\bv, \bvp \in \R^n$, the coupled-LDOS histogram is a $B$-dimensional vector, denoted $\hcldos_{\bv,\bvp} \in \R^B$, where 
	\vspace{-0.05in}
	\beq
	\label{eq:cldosfeat}
\hspace{-0.15in}	\hcldos_{\bv,\bvp}(\widetilde{\lambda}_b) = \frac{1}{\Delta} \frac{\sum_{i=1}^n (\bv^T \bu_i) (\bu_i^T\bvp)\; \I(\lambda_i \in \text{\em Bin}(\widetilde{\lambda}_b))}{n} \;, \forall b 
	\vspace{-0.05in}
	\eeq
\end{definition}

{Note that $(\bv^T \bu_i) (\bu_i^T\bvp)$ is the $i$-th entry of $\widehat{\bc}_{\bv,\bvp}$ (from Eq. \ref{eq:conv}).
Hence $\hcldos_{\bv,\bvp}$ can simply be viewed as $\widehat{\bc}_{\bv,\bvp}$, binned according to the corresponding eigenvalues of each entry.}

Moreover, recall that we use the GQL algorithm to approximate the LDOS features, where the terms $\bv^T \bu_i$ or $\bu_i^T\bvp$ are not computed using the individual eigenvectors explicitly. Nevertheless it is easy to acquire  cLDOS features in Eq. \eqref{eq:cldosfeat} using the LDOS features in Eq. \ref{eq:ldosfeat} and simple algebra. 
Given the separate LDOS features for $\bv$ and $\bvp$, we also create those for $(\bv+\bvp)$.
Then,
\vspace{-0.05in}
\beq
\hcldos_{\bv,\bvp} = [\hldos_{\bv+\bvp} - \hldos_{\bv} - \hldos_{\bvp}]/2.
\label{eq:cldos_arithmetic}
\vspace{-0.05in}
\eeq

\hide{
————————————————————
Handling various attributes
————————————————————

*) Categorical: 2 cases
Case 1: when set of unique values differ across graphs;  e.g. high-school
for each unique value val:
1-hot encode as u and compute LDOS-integral ( i.e. $u^T.f(A).u$ )
output histogram of LDOS values across all unique values
(the histogram makes comparable across graphs)

Case 2: when set of unique values is same across graphs;  e.g. gender

for each unique value attribute\_val:
1-hot encode as u and compute LDOS-integral  ( i.e. $u^T.f(A).u$ )
report as individual graph feature for $< attribute_name, attribute_val>$
(so LDOS of <gender, female> and <gender, male> will be TWO separate graph-features)

*) Numerical: remember to zero-mean the vector before feeding to LDOS

*) In addition we can try using node-level *structural* features as u, e.g. u[v] = degree(v) or some centrality(v) like Pagerank, eccentricity (longest shortest path to all nodes), local clustering coefficient.

Append all features above across all attributes to obtain the LDOS-profile of a graph.

NOTE: LDOS (the histogram of locally-adjusted eigenvalues) alone is a set of frequency-features, but for starters, we can keep it aside.
}

\vspace{-0.05in}

\subsection{Functions over the Spectrum}

DOS, LDOS and cLDOS histograms provide ``raw'' information about the graph spectrum and the attributes.
In addition, we define aggregate scalar features by specifying various frequency response functions (\frf) \cite{balcilar2021analyzing} (Eq. \eqref{eq:frf}) over these histograms. 
\begin{definition}[(cL)DOS aggregate features]
	Given a DOS, LDOS or cLDOS histogram $\bh \in \R^B$, and a frequency response func. $\phi(\cdot)$, a (cL)DOS aggregate feature $g_{\phi} \in \R$ is written as
	\vspace{-0.05in}
	\beq
	\label{eq:agg}
		g_{\phi} =	\sum_{b=1}^B \;  \bh(\widetilde{\lambda}_b) \; \phi(\widetilde{\lambda}_b) 
	\vspace{-0.05in}
	\eeq
\end{definition}
Each \frf $\phi(\cdot)$ focuses on a different part of the spectrum, inducing a variety of graph filters.
In Fig. \ref{fig:ldos} (bottom) we show three example FRFs; a low-pass one (blue) that has high values for smaller eigenvalues, a mid-pass one (red) as well as one that is both low-and-high pass (orange).
To extract graph connectivity and attribute information broadly, 
we construct a \textit{portfolio} 
of these graph filters, i.e. associated \frf's $\{\phi(\cdot)\}$, called a \textbf{filterbank}.


Before delving into the details of our filterbank, we make a few remarks.
First, note that the sum in Eq. \eqref{eq:agg} is an approximation of the integral in Eq. \eqref{eq:dosint} for $\hdos$,
that of Eq. \eqref{eq:ldosint} for $\hldos$, and accordingly an approximation of $\bv^T \phi(\bS) \bvp$ for $\hcldos$.
Second, given the efficiently-computed histograms thanks to the GQL algorithm, computing the aggregate features by Eq. \eqref{eq:agg} is extremely fast and simply involves a weighted sum.
This allows us to employ a large filterbank containing many different \frf's almost for ``free''.
Finally, we have seen that our cLDOS aggregate features relate to graph signal convolution.
Denoting the vector of frequency responses by 
$
\bphi := \{ \phi(\lambda_i) \}_{i=1}^n
$, 
based on Eq.s \eqref{eq:ldosint} and \eqref{eq:conv}, 
\vspace{-0.05in}
\beq
\hspace{-0.15in}
\int \phi(\lambda) f(\lambda; \bv, \bvp) 
=  \sum_{i=1}^n \phi(\lambda_i) (\bv^T \bu_i) (\bu_i^T \bv) 
= \bphi^T \widehat{\bc}_{\bv,\bvp} 
\vspace{-0.05in}
\eeq

In the following, we present two classes of FRF's that \method employs to extract (cL)DOS aggregate features.

\subsubsection{\bf Chebyshev polynomials}

We use the series of Chebyshev polynomials as a set of \frf's defined by the recurrence $\phi_1(\lambda) = 1, \; \phi_2(\lambda) = 2({\lambda}/{\lambda_{\max}}) -1$, and
 $\phi_k(\lambda)  = 2 \phi_2(\lambda) \phi_{k-1}(\lambda) - \phi_{k-2}(\lambda)$,
where $\lambda_{\max}$ is the maximum eigenvalue.

\begin{figure}[!t]
	\vspace{-0.1in}
	\centering
	\begin{tabular}{cc}
			\includegraphics[width=0.5\linewidth]{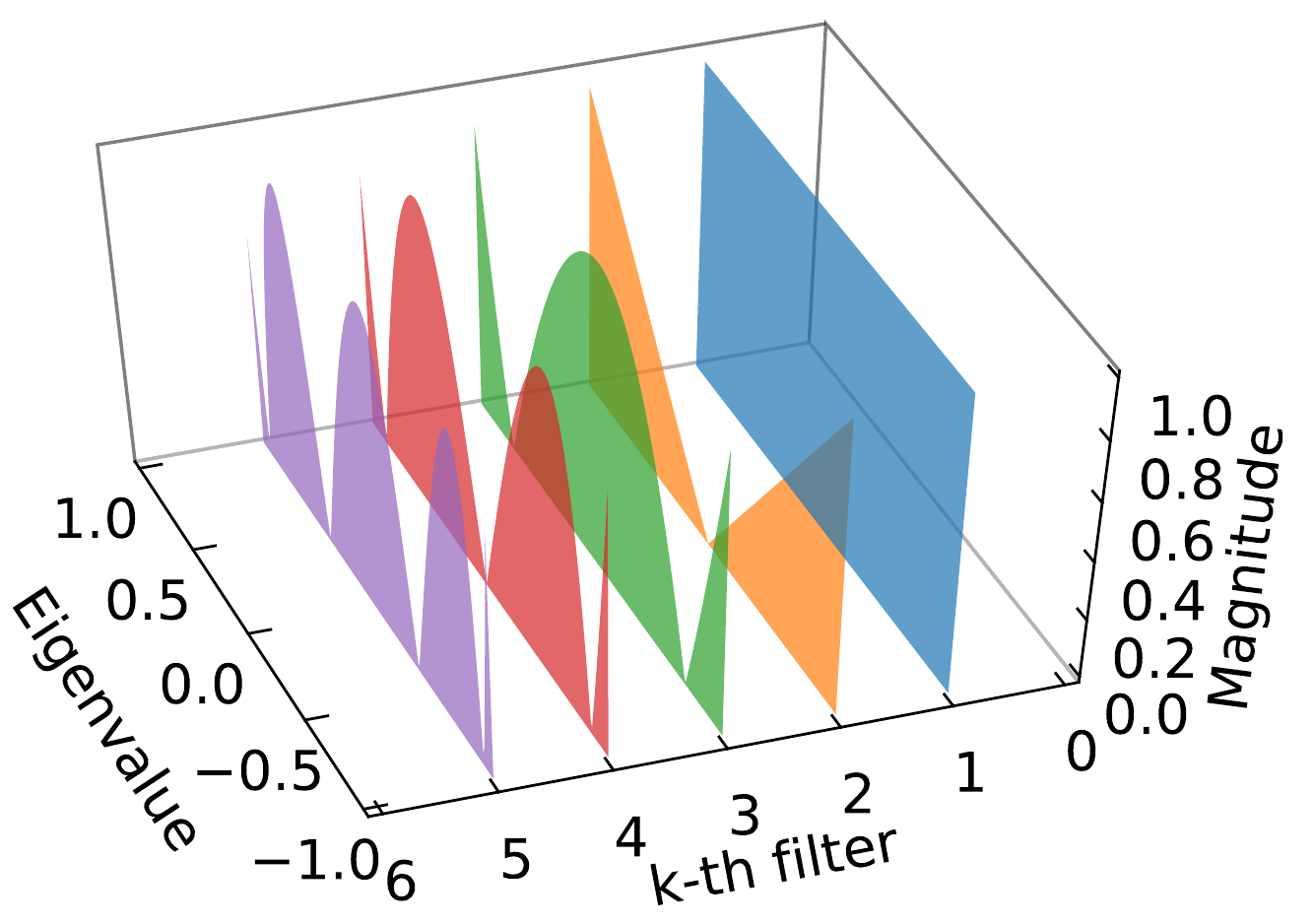} &
		\hspace{-0.2in}	
		\includegraphics[width=0.48\linewidth]{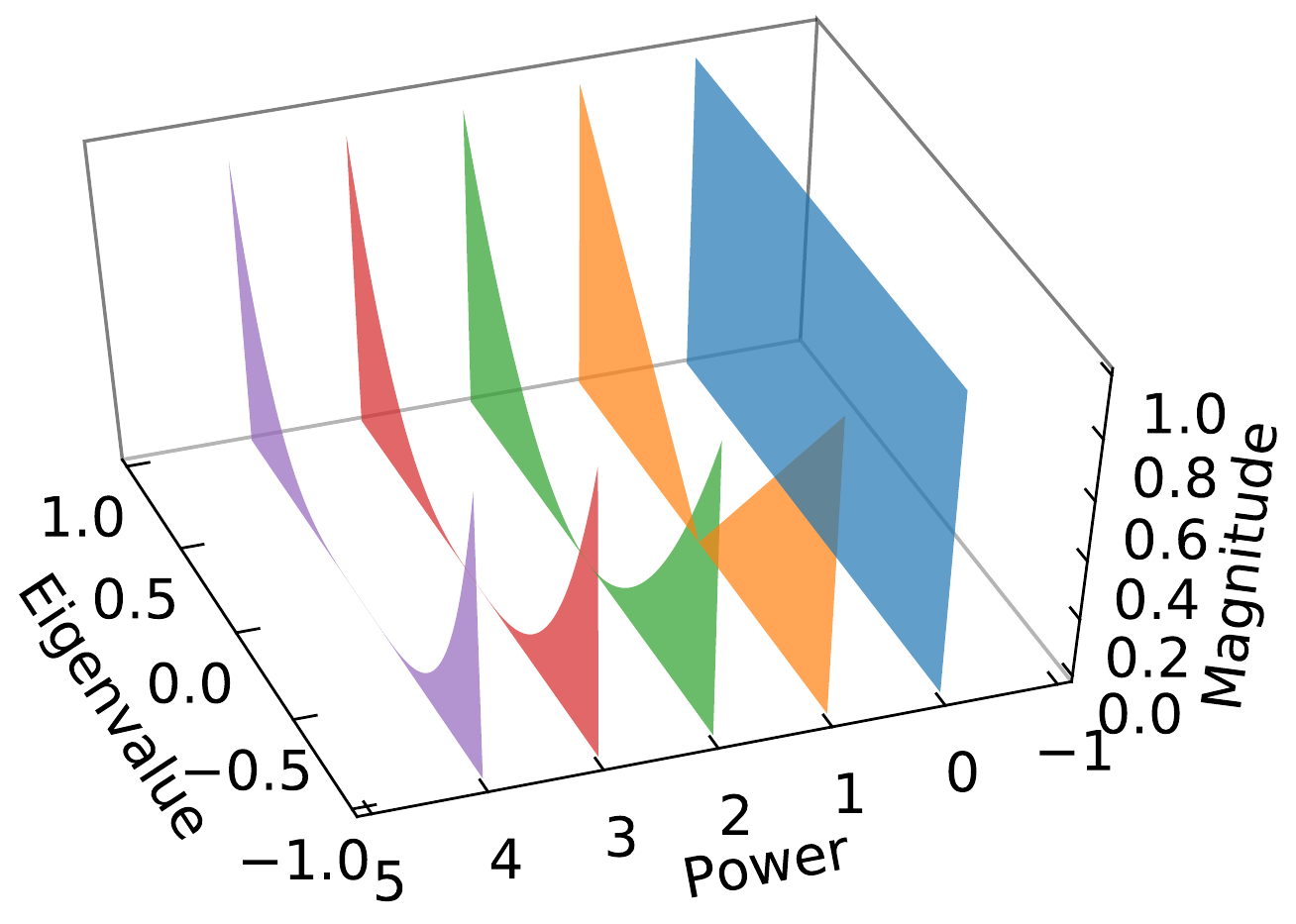} \\
		(a) Chebyshev polynomials & (b) Power functions 
	\end{tabular}
	\vspace{-0.1in}
	\caption{Frequency response functions $\phi_k(\lambda)$ (magnitude in absolute val.s) \label{fig:filterbank}}
	\vspace{-0.1in}
\end{figure}

{\bf Interpretation.} As shown in Fig. \ref{fig:filterbank}(a), Chebyshev polynomials provide frequency profiles that cover various parts of the spectrum. For example, the 2nd one is mostly a
low- and high-pass filter and stops the middle band, while the 3rd one passes the middle bands as well as very high and very low bands of the spectrum, and so on. Given a number of these \frf's, emphasis can be put on passing/stopping specific bands by a weighted combination of them. 

The flexibility of any-band filtering by \method is favorable over several modern graph neural networks (GNNs).
GCN \cite{Kipf2016tc}, for instance, works as a low-pass-only filter and hence does not cover the whole spectrum.
GIN \cite{Xu2019ty} has a learnable scalar parameter $\epsilon$ that determines which band to stop, however its \frf is a
linearly decreasing function, which is not a strong low-pass or high-pass filter. (See Fig. 2 in \cite{balcilar2021analyzing}.)
In contrast, spectrally-designed ChebNet \cite{conf/nips/DefferrardBV16} is more expressive and also employs the Chebyshev polynomials.
We compare to these modern GNNs in the experiments on graph classification tasks.

\subsubsection{\bf  Power functions}
The second class of \frf's in our filterbank uses (both positive and negative) powers of the spectrum, that is,
\vspace{-0.05in}
\begin{align*}
\phi_k(\lambda) = \lambda^k \;,\; k=\pm \{1, \ldots, K/2\}
\vspace{-0.05in}
\end{align*}


{\bf Interpretation.~} Our aggregate features using the power functions relate to random walks on the graph.
Consider positive values of $k$ and $\bS = \tW$.
Recall that for $\hdos$, Eq. \eqref{eq:agg} is an approximation of $\text{trace}(\phi(\bS))$, which is equal to the total return-probability of a $k$-step random walk to a node.
For $\hldos$, aggregate features approximate $\bv^T \phi(\bS) \bv$. For a binary/categorical attribute where $\bv$ depicts a certain value, e.g. $val$ $:=$ ({\tt party\_affiliation:democrat}), it corresponds to the probability that a $k$-step random walk starting at any node with value $val$  ``hits''/reaches another node with the same value. 
For $\hcldos$, similarly, it is the probability that such a walk starting at any node with a certain $val$ will reach another node with a different $val^\prime$.
Moreover, for two continuous attributes $\bv$ and $\bvp$, approximating $\bv^T \phi(\bS) \bvp$ via $\hcldos$ would capture the covariance of the attributes over ``$k$-hop connected'' pairs of nodes that can reach each other within $k$-steps. 

The interpretations extend to the negative powers as well,
which correspond to many walks of different lengths in the limit.
In that respect, aggregate features using power functions depict \textit{multi-scale} properties, where increasingly positive values of $k$ capture microscopic to mesoscopic properties related to short/local random walks, whereas
negative powers relate to the long-range walks and thereby macroscopic structure.








\begin{table}[!t]
	\caption{Summary of \method graph-level features based on spectral densities (cL)DOS, organized as histogram and aggregate features using Chebyshev polynomials and power functions.   \label{tab:features}}
	\vspace{-0.1in}
	\centering
	\begin{tabular}{|p{0.15in}|p{0.2in}|p{0.2in}|p{0.2in}|p{0.2in}|p{0.2in}|p{0.25in}|p{0.25in}|p{0.25in}|}
		\hline
		\multicolumn{3}{|c|}{\textbf{DOS}}                   & \multicolumn{3}{c|}{\textbf{LDOS}}                   & \multicolumn{3}{c|}{\textbf{cLDOS}}\Tstrut\Bstrut                    \\
		\hline
			\multirow{ 2}{*}{$\bh$ist} & \multicolumn{2}{c|}{agg. $g_{\phi}$} & \multirow{ 2}{*}{$\bh$ist}  & \multicolumn{2}{c|}{agg. $g_{\phi}$} & \multirow{ 2}{*}{$\bh$ist}   & \multicolumn{2}{c|}{agg. $g_{\phi}$}\Tstrut\Bstrut \\
			\cline{2-3}\cline{5-6}\cline{8-9}
		& Cheb.         & Pow.         &            & Cheb.         & Pow.        &              & Cheb.         & Pow.\Tstrut\Bstrut         \\
		\hline
		$B$         &    $K$           &       $K$        & $B D$ &      $KD$         &     $KD$          & $B {D \choose 2}$ &     $K {D \choose 2}$          &  $K {D \choose 2}\Tstrut\Bstrut$  \\
		\hline          
	\end{tabular}
\vspace{-0.2in}
\end{table}

We conclude with an overview of all the graph-level features described in this section.
Table \ref{tab:features} gives the number of features by category, where $B$ is the number of histogram bins,
$K$ is the number of Chebyshev/power \frf's, and $D \geq d$ is the total number of attributes upon one-hot-encoding the categorical and binary attributes. \method yields 
$(B+2K)(1 + D + {D \choose 2})$ features in total for an attributed graph, which are permutation- and size-invariant, task-agnostic, 
variable band-pass, multi-scale, and extremely efficient to compute.
We outline the steps in Fig. \ref{fig:algo} and give detailed complexity analysis next.

\begin{figure}
\fbox{\begin{minipage}{\linewidth}
\textbf{Input:} Graph $G$ (with $D$ node attributes), parameters $B$, $K$.
\begin{itemize}[leftmargin=*, itemsep=0pt]
	\item (Preprocess) Compute $2K$ agg. functions on $B$ bin centers.
	\item $\tW \leftarrow$ normalized adjacency matrix of $G$
	\item $\bv_d \leftarrow$ attribute vector for each $d \in \{1\ldots D\}$
	\item Compute $\hdos$ using GQL \cite{dong2019network} on $\tW$
	\item For each $d$, compute $\hldos_{\bv_d}$ using GQL on $\tW, \bv_d$.
	\item For each pair $d, d' \in \{1\ldots D\}$, compute $\hldos_{\bv_{d}+\bv_{d'}}$ using GQL and then $\hcldos_{\bv_{d},\bv_{d'}}$ using Eq. \eqref{eq:cldos_arithmetic}.
	\item For each histogram computed above, dot product with all aggregate functions to produce aggregate features.
\end{itemize}







\end{minipage}}
\caption{Steps to generate all \method features (See Table \ref{tab:features}) \label{fig:algo}}
\vspace{-0.2in}
\end{figure}

\subsection{Computational Complexity}

Since \method computes an embedding for each graph independently,
it scales linearly with the number of graphs in the dataset, i.e., $N$.

We analyze the asymptotic runtime of \method on a single graph $G$ with $n$ nodes, $m$ edges, and $D$ total node attributes (including one-hot encoded labels and categorical attributes).
We use the Gauss Quadrature and Lanczos  algorithm described by Dong {\em et al.} \cite{dong2019network} to compute a (cL)DOS \textit{histogram}. This involves ($i$) running $\eta_L$ Lanczos iterations, each requiring $O(m)$ operations, followed by ($ii$) the eigendecomposition of a tridiagonal $n$$\times$$n$ matrix, with $O(n^2)$ operations.
Note that although a tridiagonal matrix eigendecomposition has a quadratic worst-case complexity theoretically, this operation is extremely fast in practice -- especially for real-world matrices.
 Each \textit{aggregate feature} 
 requires a dot product of two vectors of size $B$ for $O(B)$, 
  where we use $2K$ different frequency response functions (i.e. $\phi(\cdot)$'s) in total ($K$ each for Chebyshev and powers). 
  Then, the total complexity of computing one histogram and its related aggregate features is $O(n^2 + \eta_L m + KB)$.
This gives a total runtime of $O\left((n^2 + \eta_L m + KB) \cdot \alpha\right)$, where $\alpha$ denotes the number of desired graph-level features (i.e. embedding size) in \method. 

Notably, \method is modular and can include any subset of the features in Table \ref{tab:features}.
For datasets with a large number of node attributes, one can skip cLDOS features, or only choose important attribute-pairs to ensure $\alpha = O(D)$.
Also note that each aggregate feature for a given $\phi(\cdot)$ can be computed independently, and hence can be easily parallelized.


\section{Experiments}
\label{sec:experiments}

To evaluate \method we design both quantitative and qualitative experiments to answer the following questions.
\bit
\item[\textbf{Q1. Graph Classification}~] How does \method (unsupervised) compare to the modern GNNs and graph kernels (un/supervised) on benchmark graph classification tasks?
\item[\textbf{Q2. Exploratory Graph Analysis}]  Can \method provide insights for mining real-world attributed graphs?  
\item[\textbf{Q3. Efficiency}~] How fast and scalable is \method?
\eit

\subsection{Experiment Setup}
\label{ssec:setup}

\textbf{Datasets.~} List of datasets and summary are in Table \ref{tab:data}.

\hide{
	%
	%
	%
	%
	%
	%
	%
	%
	%
	
}


For graph classification, we use eight benchmark datasets from TUDataset repository\footnote{\url{https://chrsmrrs.github.io/datasets/docs/datasets/}}.
Five are commonly used social network datasets, \redditb, \reddit, \collab, \imdbb and \imdb.
These contain only plain graphs--a setting with which all the baselines are compatible.
The other three are biochemistry datasets, \protein, \dd and \aids, which have node labels and/or attributes.

We also use four other graph datasets to specifically showcase the strengths of \method in leveraging the full graph spectrum.

%
%
%
%
%
%

$\sbt$ \band is a synthetic dataset consisting of images generated via sinusoidal patterns from two frequency ranges \cite{balcilar2021analyzing}.
$\sbt$ \congress is based on the voting patterns in 41 U.S. Senates (1927--2008)
\cite{cucuringu2011localization}, where nodes represent senators (labeled by party affiliation) and edge weights represent voting agreement.
To create separate classes of graphs, we add noise to edge weights between same-party senators (class 1), 
and randomly picked senators (class 2) in one randomly picked congress. 
$\sbt$ \congressl is generated by picking one congress at random and shuffling the labels of senators via random swaps; 50 swaps in class 1, and 300 in class 2.
$\sbt$ \mig  is based on the county-to-county migration in the U.S. \cite{cucuringu2011localization}.
To create separate classes of graphs, we add noise to edges between a pair of bordering states (class 1) or within a state (class 2).

In addition to the above datasets, we perform graph exploratory analysis using \method on two more datasets:

$\sbt$ \face consists of Facebook college social networks from 100 American institutions \cite{traud2012facebook}, with student demographic information (major, dorm, status, class-year, etc.) as node attributes.
$\sbt$ \states  is built from the \mig dataset, by inducing 49 separate graphs - one for each mainland state and its bordering states. We label counties of the selected state as 1, and the counties of the neighbors as 2.

\textbf{Baselines.~} 
We compare \method quantitatively to various unsupervised and supervised graph embedding, graph kernel, and graph neural network methods on graph classification tasks.

\textit{Unsupervised explicit graph embeddings} are in the same category as \method and hence most comparable.
As baselines from this category, we use $\sbt$ \fgsd~\cite{verma2017hunt}, $\sbt$ \netlsd~\cite{tsitsulin2018netlsd} and $\sbt$ \gvec~\cite{narayanan2017graph2vec}, which we described briefly in \S\ref{ssec:prior}.

%
%

\textit{Graph kernels} are also unsupervised; here we use three of the best performing kernels on classification benchmarks, and a recent DOS-based graph kernel.
$\sbt$ \wl \cite{journals/jmlr/ShervashidzeSLMB11}: the Weisfeiler-Lehman graph kernel,
$\sbt$ \wloa \cite{conf/nips/KriegeGW16}: the Weisfeiler-Lehman Optimal Assignment kernel,
$\sbt$ \pk \cite{journals/ml/NeumannGBK16}: the Propagation Kernel, and
$\sbt$ \dos \cite{huang2021density}, the Density of States Graph Kernel.

\textit{GNN baselines} include state-of-the-art supervised models, such as
$\sbt$ \cheb \cite{conf/nips/DefferrardBV16},
$\sbt$ \gcn \cite{Kipf2016tc}, and 
$\sbt$ \gin \cite{Xu2019ty}. 

Note that \fgsd, \netlsd, and \dos are for plain graphs only. 
\gvec, \wl, and \wloa admit node labels but not (continuous) attributes.
Therefore, they input only the admissible parts of a graph dataset for classification.

\textbf{Model configuration.~} 
In our experiments with \method, we set $\eta_L$$=$$100$, 
$B$$=$$200$ and $K$$=$$100$ (see Table \ref{tab:features}).
For plain datasets, we use node degree as a continuous attribute.
For \fgsd, we use
$L^{-1}$ as the distance function
and $0.001$ as the binwidth.
For \netlsd,
we use heat trace signatures at 250 different values of $t$ logarithmically spaced in $[10^{-2}, 10^2]$.
For \gvec,
we set the WL iteration count to $5$ and output dimension to 1024.
For the kernels \wl, \wloa and \pk, we use the implemention from the GraKel package\footnote{\url{https://ysig.github.io/GraKeL/}}, and the default parameters suggested. 
For \dos, same as with \method, we use 200 bins and 100 Chebyshev moments.
For all the GNNs, we use mean-pooling as the readout function.

We run all non-GNN experiments on one core of Intel(R) Xeon(R) CPU E5-2667 v3 CPU @3.20GHz.
GNN experiments are run on a server with NVIDIA Tesla V100 GPU and one core of Intel(R) Xeon(R) Gold 6248 CPU @2.50GHz.
\vspace{-0.15in}

\hide{
	%
	%
	%
	%
	%
}

\hspace{-0.3in}
\begin{table}[]
	
	\centering
	\caption{Dataset summary statistics.}
	\vspace{-0.05in}
	\begin{tabular}{l|rp{0.1in}rrrr}
		\hline
		          & $N$ & Cls. & Avg. $n$ &  Avg. $m$ & Lbl. & Attr.\Tstrut\Bstrut \\ \hline
		\redditb  &   2000 &     2 &   429.6 &    497.7 & -      & -\Tstrut     \\
		\reddit   &   5000 &     5 &   508.5 &    594.8 & -      & -     \\
		\collab   &   5000 &     3 &    74.5 &   2457.8 & -      & -     \\
		\imdbb    &   1000 &     2 &    19.8 &     96.5 & -      & -     \\
		\imdb     &   1500 &     3 &       13.0 &     65.9 & -      & -     \\
		\dd       &   1178 &     2 &   284.3 &    715.7 & 78     & -     \\
		\protein  &   1113 &     2 &    39.1 &     72.8 & -      & 1     \\
		\aids     &   2000 &     2 &    15.7 &      16.2 & 38     & 4\Bstrut     \\ \hline
		\band     &   5000 &     2 &      200 &    1072.6 & -      & 1 \Tstrut    \\
		\congress &    200 &     2 &     4196 &  450662.5 & 3      & -     \\
		\mig      &    200 &     2 &     3075 &   1092282.0 & 19     & - \Bstrut    \\ \hline
		\face     &    100 &   n/a & 12083.2 & 469845.4 & -      & 7 \Tstrut    \\
		\states   &     49 &   n/a &    367.4 &  21633.9 & 2      & -\Bstrut     \\ \hline
	\end{tabular}
\label{tab:data}
\vspace{-0.2in}
\end{table}

\setlength{\tabcolsep}{3pt}

\begin{table*}[!t]
	\centering
		\caption{Graph classification performance by \method and its DOS-only (i.e. no attributes) variant {\sc DOGE}, compared with three types of baselines. The highest performance per dataset is {\BFSERIES in bold} and the highest among unsupervised methods is \underline{underlined}.
		E denotes the code outputting an error; The numbers with symbols denote the paper from which the numbers are taken: \fnddag \cite{journals/ans/KriegeJM20}, $^\ast$\cite{huang2021density}, \fndag \cite{balcilar2021analyzing}.
	\vspace{-0.1in}
}
	\begin{tabular}{l|ll|lll|llll|lll}
		\hline
		                &                                                  \multicolumn{5}{c|}{Graph Embedding (Unsupervised)}                                                  &                                      \multicolumn{4}{c|}{Graph Kernels (Unsupervised)}                                       &                      \multicolumn{3}{c}{GNNs (Supervised)}\Tstrut\Bstrut                      \\ \hline
		                & \multicolumn{1}{c}{\method}        & \multicolumn{1}{c|}{\doge} & \multicolumn{1}{c}{\fgsd} & \multicolumn{1}{c}{\netlsd} & \multicolumn{1}{c|}{\gvec} & \multicolumn{1}{c}{\wl} & \multicolumn{1}{c}{\wloa}     & \multicolumn{1}{c}{\pk}            & \multicolumn{1}{c|}{\dos}     & \multicolumn{1}{c}{\cheb} & \multicolumn{1}{c}{\gcn} & \multicolumn{1}{c}{\gin}\Tstrut\Bstrut \\ \hline
		\texttt{RED-B}  & \underline{91.6} (1.5)             & 90.3 (1.8)                & 82.4 (2.6)                & 85.6 (2.2)                  & 74.2 (2.7)                 & 83.9 (0.5)\fnddag       & 88.9 (0.1)\fnddag             & 85.5 (0.3)\fnddag                  & 88.8 (0.3)$^\ast$             & 90.2 (2.0)                & 89.9 (2.0)               & {\BFSERIES 91.7} (1.6)\Tstrut   \\
		\texttt{RED-5K} & {\BFSERIES \underline{55.6}} (2.2) & 53.8 (2.1)                & 47.0 (1.8)                & 45.9 (2.1)                  & 41.5 (1.6)                 & 51.2 (0.3)$^\ast$       & E                             & E                                  & 52.8 (0.2)$^\ast$             & {55.0} (2.2)    & 54.2 (1.7)               & 54.7 (2.0)               \\
		\collab         & 72.2 (2.0)                         & 72.2 (2.0)                & 70.2 (1.8)                & 68.4 (1.9)                  & 57.9 (1.5)                 & 74.8 (0.2)$^\ast$       & 79.8 (1.6)                    & 77.8 (1.7)                         & \underline{80.8} (0.2)$^\ast$ & {\BFSERIES 84.6} (1.1)    & 84.2 (1.2)               & 83.8 (1.6)               \\
		\texttt{IMDB-B} & 72.6 (4.3)                         & 71.6 (4.3)                & 70.6 (4.1)                & 69.7 (4.1)                  & 56.0 (4.1)                 & 71.3 (1.0)\fnddag       & \underline{73.5} (0.6)        & 71.2 (0.7)\fnddag                  & 72.8 (0.9)$^\ast$             & 80.2 (3.9)                & 79.9 (3.7)               & {\BFSERIES 80.8} (4.5)   \\
		\texttt{IMDB-M} & 47.8 (3.5)                         & 47.6 (3.7)                & 48.6 (3.4)                & 47.9 (3.7)                  & 44.4 (3.8)                 & 50.7 (0.6)\fnddag       & 50.7 (0.5)\fnddag             & \underline{51.0} (0.7)\fnddag      & 49.4 (0.5)$^\ast$             & 55.6 (2.7)                & 55.2 (2.7)               & {\BFSERIES 56.3} (3.1)   \\
		\dd             & 80.1 (3.5)                         & 76.2 (3.4)                & 76.5 (3.5)                & 76.6 (3.5)                  & 76.2 (3.5)                 & 80.9 (0.3)              & 79.9 (0.5)                    & {\BFSERIES \underline{81.6}} (0.5) & 73.4 (3.7)                    & 78.9 (1.9)                & 78.0 (1.8)               & 79.3 (1.9)               \\
		\texttt{PROTN}  & 74.9 (3.5)                         & 74.9 (3.5)                & 74.2 (3.3)                & 74.5 (4.0)                  & 72.1 (3.1)                 & 73.9 (0.7)\fnddag       & \underline{75.9} (0.6)\fnddag & 74.6 (0.5)\fnddag                  & 72.1 (3.9)                    & 78.3 (2.7)                & 76.7 (3.5)               & {\BFSERIES 78.4} (3.9)   \\
		\aids           & {\BFSERIES \underline{99.8}} (0.3) & 99.8 (0.3)                & 99.6 (0.4)                & 99.6 (0.5)                  & 98.8 (0.7)                 & 99.7 (0.0)\fnddag       & 99.7 (0.0)\fnddag             & 99.7 (0.0)\fnddag                  & 99.1 (0.7)                    & 96.9 (1.6)                & 95.5 (1.3)               & 98.6 (0.6)\Bstrut               \\ \hline
		\texttt{Cong}   & {\BFSERIES \underline{99.5}} (1.5) & 54.7 (11.0)               & 95.1 (4.3)                & 99.5 (1.5)                  & 86.8 (7.4)                 & 84.8 (7.3)              & 81.1 (7.7)                    & 68.6 (8.3)                         & 60.0 (10)                     & 50.0 (0.0)                & 50.0 (0.0)               & 57.0 (5.9)\Tstrut                         \\
		\texttt{Cong-l} & {\BFSERIES \underline{78.0}} (8.6) & 58.9 (10.0)               & 50.0 (0.0)                & 60.4 (9.7)                  & 59.8 (11)                  & 62.2 (10)               & 62.3 (10)                     & 58.2 (10)                          & 55.7 (9.7)                    & 50.0 (0.0)                & 50.0 (0.0)               & 71.5 (9.4)               \\
		\mig            & {\BFSERIES \underline{100.0}} (0)  & 62.3 (9.7)                & 99.5 (1.5)                & 99.9 (1.1)                  & 50.0 (0.0)                 & 99.8 (1.4)              & 99.8 (1.4)                    & 100 (0.0)                          & 53.5 (12)                     & 100.0 (0.0)                & 78.5 (1.7)               & 100.0 (0.0)               \\
		\texttt{BPass}  & \underline{90.8}                   & 51.9                      & 47.9                      & 51.4                        & 50                         & 50                      & 51.6                          & 70.4                               & 48.5                          & {\BFSERIES 98.2}\fndag    & 77.9\fndag               & 87.6\fndag\Bstrut               \\ \hline
		Avg.            & 82.5                               & 69.1                      & 74.1                      & 75.8                        & 66.0                       & 75.6                    & 76.6                          & 76.3                               & 68.6                          &        78.4                   &       74.2                   &          80.5\Tstrut\Bstrut                \\ \hline
	\end{tabular}
\label{tab:class}
	\vspace{-0.1in}
\end{table*}

\subsection{Graph Classification}
\label{ssec:class}

\textbf{Classifier configurations.~}
For classification with the embeddings produced by unsupervised methods, we use the kernel-SVM\footnote{SVM facilitates comparable results between implicit and explicit kernels.} classifier 
with the regularization parameter $C$ chosen from $\{10^{-3}, 10^{-2}, \ldots, 10^3\}$ via $10$-fold cross-validation.
We perform this experiment 10 times using random splits.
For explicit embeddings, we normalize each feature, and set $\gamma$ to be the inverse of the median of pairwise $\ell_2$ distances between all embeddings.
For \method, we also set the option of using LDOS, cLDOS features, and the option of using aggregate FRFs as hyperparameters. 
We normalize all kernels symmetrically.
For GNNs, we train them end-to-end using cross-entropy loss, and hyperparameters (learning-rate at $0.005$, layers in \{2,3,5,7\}, hidden sizes from \{32,64,128\} and epochs up to 200) selected via $10$-fold cross-validation.
For each of the above methods, we report the mean test accuracy for the best choice of hyperparameters, and the corresponding standard deviation on every dataset except \band, for which we use the single train-validation-test split as specified in \cite{balcilar2021analyzing}.

\textbf{Results.~}
Table~\ref{tab:class} contains all the performance results of our classification experiments.
Among the benchmark datasets, \method 
achieves on par performance with the most competitive
unsupervised baselines and 
is often comparable to (supervised) GNNs, while being considerably more resource-frugal. 

On the other four datasets, \method significantly outperforms all baseline methods due to its ability to capture the alignment of labels and attributes with graph structure at a multi-scale level, even in databases with as few as 200 graphs.
Provided \method uses considerably lower resources in comparison with kernels and GNNs, and considering that the latter are trained end-to-end, we do not expect \method to exhibit state-of-the-art performance on every dataset. Still, \method outperforms/equals all baselines on 7 of the datasets.
Moreover, \method stands out as the top choice based on average performance across all datasets. 

\begin{figure}[!ht]
	\vspace{-0.1in}
	\centering
	\includegraphics[width=0.75\linewidth]{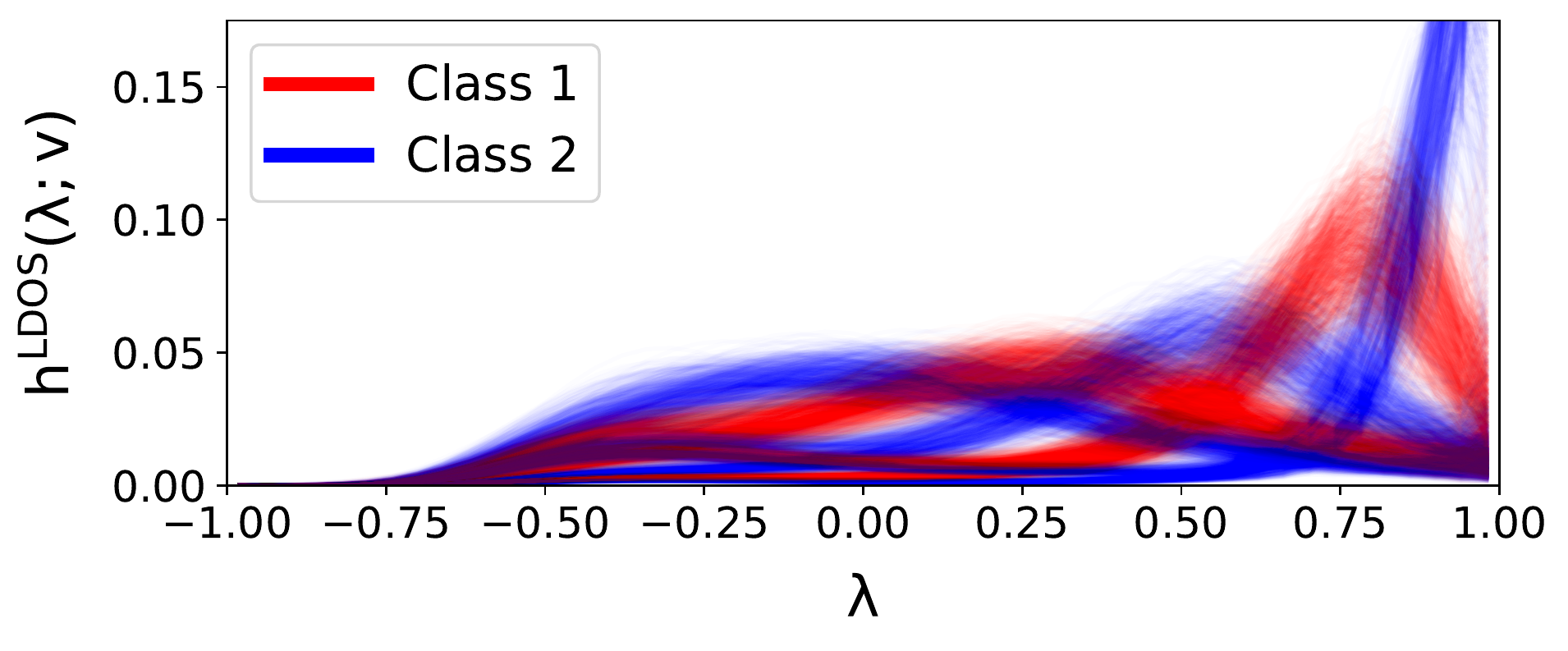}
\vspace{-0.2in}	
	\caption{\label{fig:bandpass} LDOS histograms for all graphs in \band, plotted as lines. Each class can be characterized by specific bands of eigenvalues.}
\end{figure}

On the \band dataset, only the spectrally-designed \cheb is able to outperform \method.
This can be attributed to the way that \band is created, wherein graph classes are formed based on the frequency band used to generate the underlying image. Fig.~\ref{fig:bandpass} depicts the LDOS histograms of the graphs in the \band dataset. We can clearly see that capturing specific bands of the eigenspectrum suffices to characterize the disparity between the two graph classes.

%




\textbf{Feature ablation.} 
Table~\ref{tab:class} also shows the DOS-only version of \method without using node labels and attributes, called \doge.
We observe that in the benchmark datasets, graph structure seems to hold most of the useful information needed for classification, and hence there is only a small improvement in performance from using node attributes.
In the rest of the datasets, node attributes play an important role, causing significant improvements in results for \method by using LDOS and cLDOS features.

\begin{figure}[!ht]
	\vspace{-0.1in}
	\centering
	\includegraphics[width=0.7\linewidth]{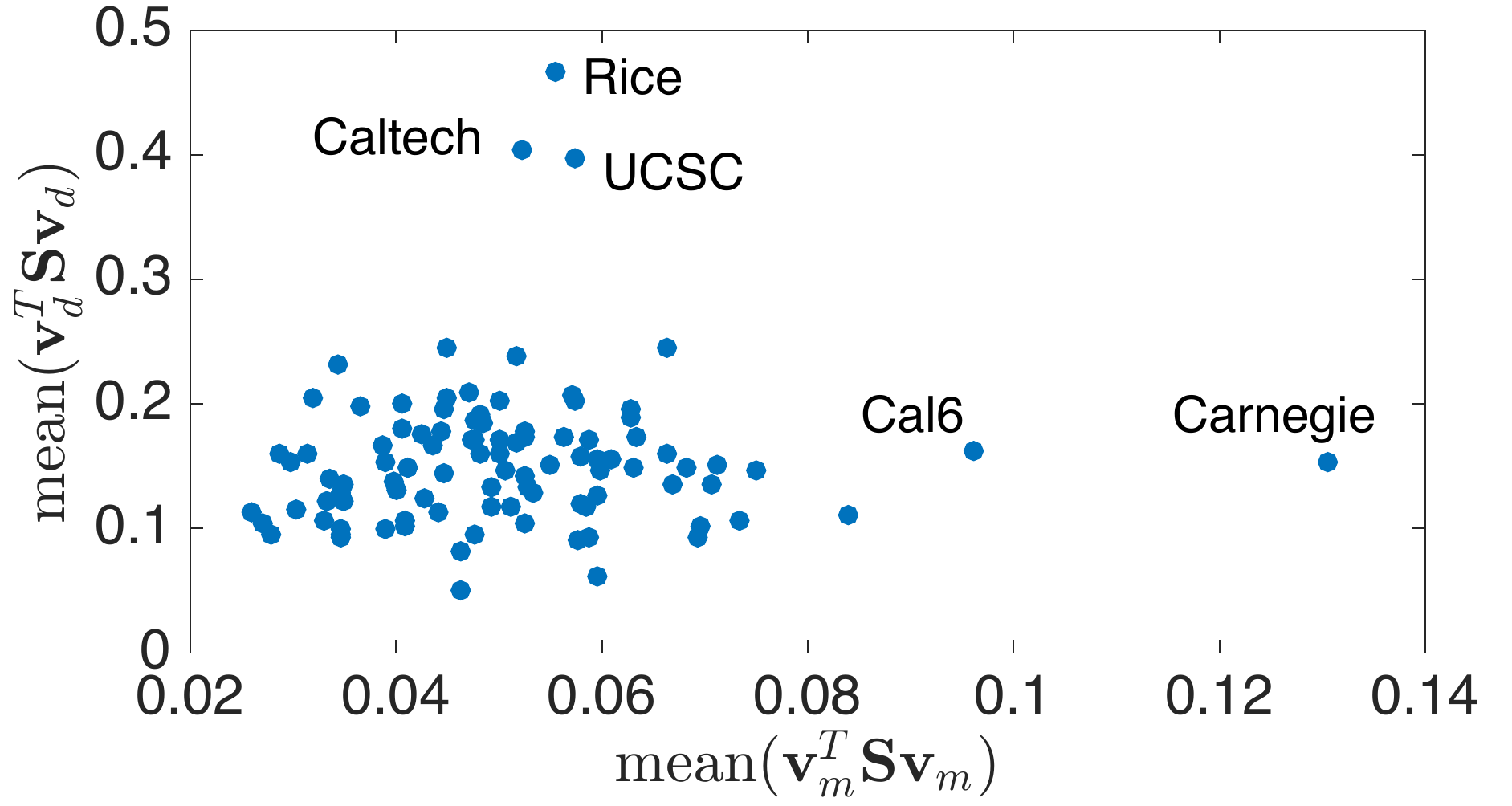}
	\vspace{-0.1in}
	\caption{\label{fig:fb_dorm_major} Average homophily w.r.t. \textit{major} vs. \textit{dorm} in 100 Facebook college social networks in the U.S., where $\bv_m$ and $\bv_d$ respectively refer to an attribute vector corresponding to a particular major $m$ and dorm $d$.}
		\vspace{-0.1in}
\end{figure}

\subsection{Graph Data Mining}

To demonstrate the interpretability of the \method features, we perform exploratory graph analysis on three real-world datasets, \face, \congress and \states.

{\underline{\face}.~} In \face, we denote each categorical feature (e.g. major) with its one-hot encoding, and hence, each particular value (e.g. Computer Science) has its own (binary) attribute vector.
We first visualize the \face graphs via {LDOS aggregate features} using these attribute vectors, with {small positive power functions} as \frf to capture the assortativity (homophily) of different attributes across different college networks.
In each graph, we compute the aggregate feature that estimates $\bv_m^T \bS \bv_m$ for every major captured by $\bv_m$, and similarly $\bv_d^T \bS \bv_d$ for every dormitory captured by $\bv_d$.
Fig.~\ref{fig:fb_dorm_major} plots the mean homophily with respect to \textit{major} and \textit{dorm} for each of the 100 colleges.

While Carnegie pops up as having the highest correlation between edges and students with the same \textit{major}, comparing the ranges of both axes suggests that \textit{dorm} is a much stronger indicator of students within a college being friends.
Moreover, this tendency seems to be more pronounced in Rice, Caltech and UCSC. This is also backed up by findings in \cite{traud2012facebook} and the real-world knowledge that Rice and Caltech are organized predominantly by dorms and other on-campus housing.

\begin{figure}[!t]
	\vspace{-0.1in}
	\centering
	\begin{tabular}{cc}
	\hspace{-0.15in}	\includegraphics[width=0.48\linewidth]{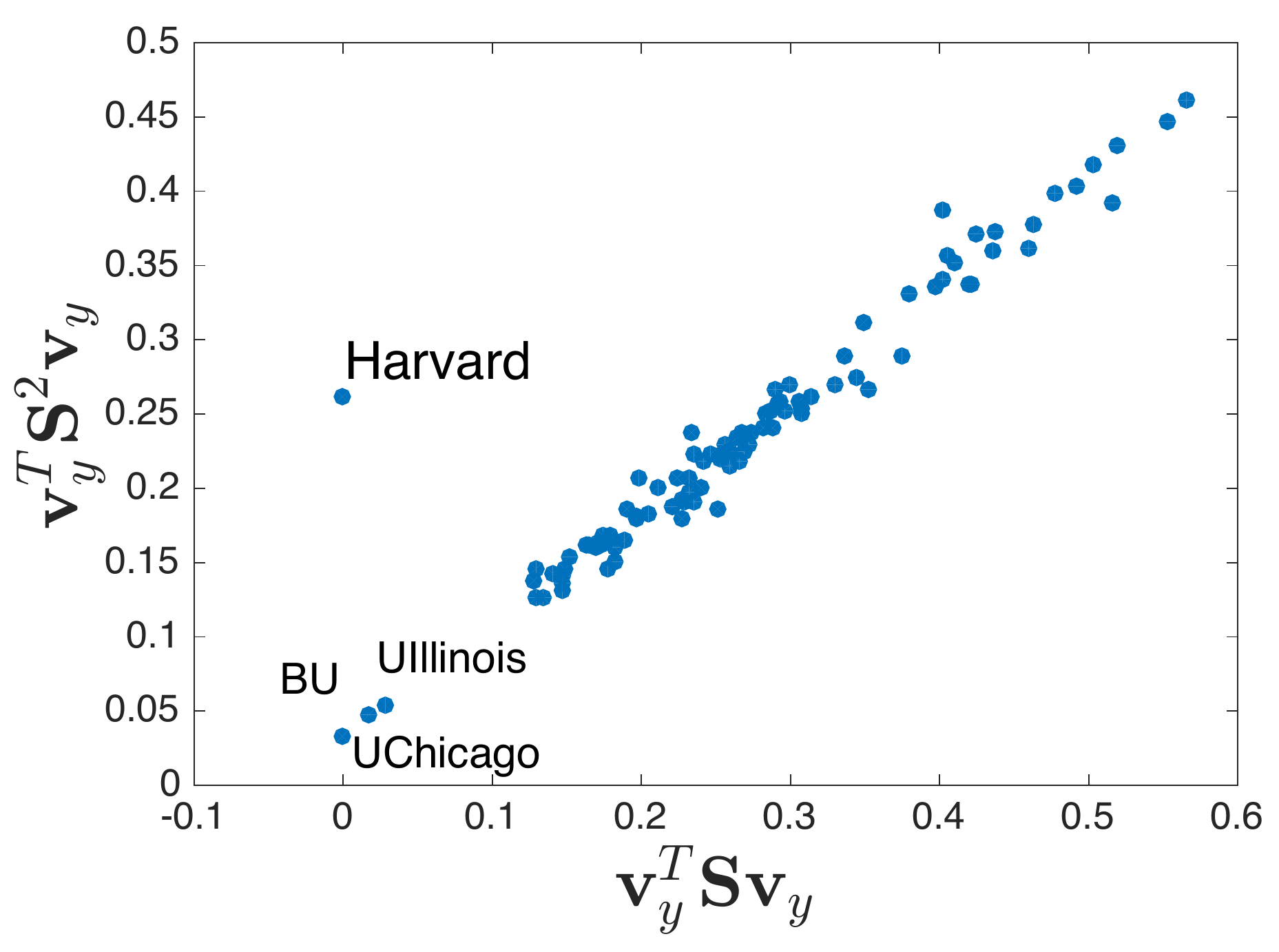}	& \hspace{-0.05in}\includegraphics[width=0.48\linewidth]{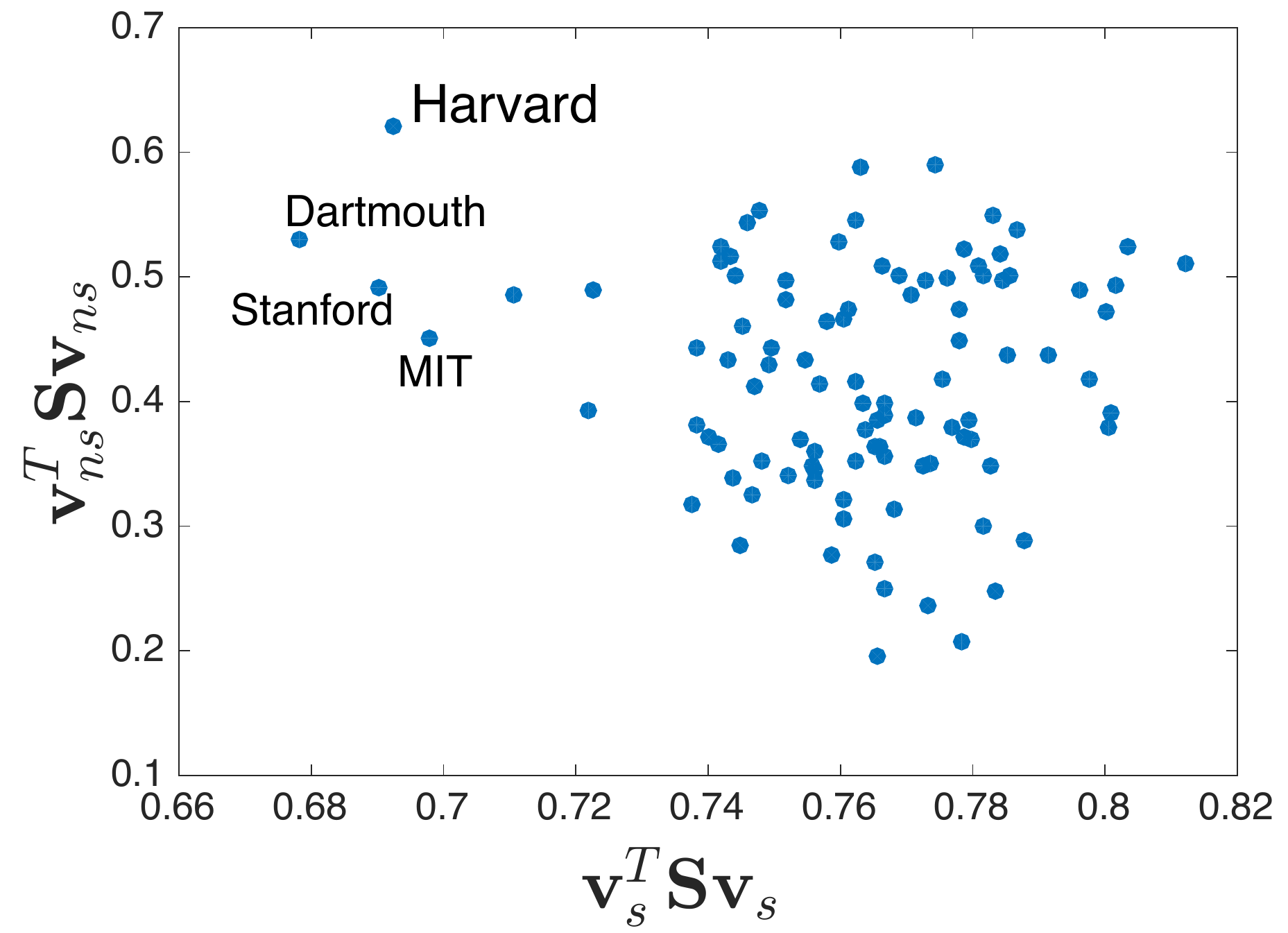} 
	\end{tabular}
	\vspace{-0.1in}
	\caption{(left) Homophily w.r.t. \textit{class\_year} based on $k$$=$$1$ and $k$$=$$2$-length paths over all 100 colleges. $\bv_y$ refers to continuous attribute vector with class years.
		(right) Homophily within student and non-student communities in all 100 colleges. Binary vector $\bv_s$ ($\bv_{ns}$) depicts student (non-student) {\em status}.\label{fig:fb_year_student}}
		\vspace{-0.2in}
\end{figure}

We also analyze similar aggregate functions over the {continuous attributes}. 
 Fig.~\ref{fig:fb_year_student}(left) plots the assortativity with respect to \textit{class\_year} for $k$$=$$1$ and $k$$=$$2$ for the power functions, which capture 1- and 2-length paths.
As we expect, these features are highly correlated in most colleges---with the striking exception of Harvard, where it appears that 2-length paths are common between individuals of similar \textit{class\_year}, but this is not the case with 1-length paths.
To investigate further, we plot homophilies for student and non-student populations for all colleges in Fig.~\ref{fig:fb_year_student}(right) and we learn that the Harvard network consists of a comparatively higher number of edges amongst non-student members, most of whom have empty or very disparate \textit{class\_year}. Even if edges between students are fewer, this is corrected when we look at 2-length paths instead.

\begin{figure}[!ht]
	\vspace{-0.15in}
	\centering
	\includegraphics[width=0.9\linewidth]{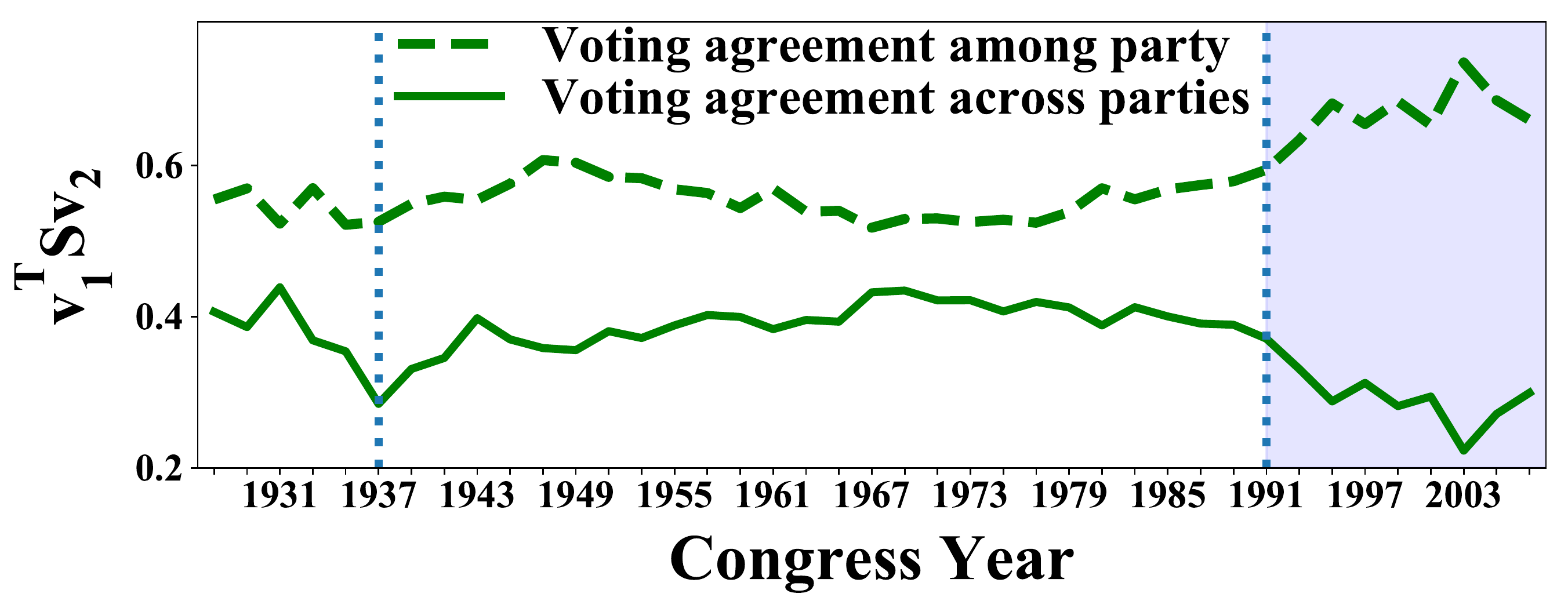}
		\vspace{-0.15in}
	\caption{\label{fig:polarization} Voting agreement within (dashed curve, $\bv_1$$=$$\bv_2$$=$$\bv_d$ or $\bv_r$) and across (solid curve, $\bv_1$$=$$\bv_d$, $\bv_2$$=$$\bv_r$) political parties over the years, for 41 Senates during 1927--2008. 
	}
\vspace{-0.1in}
\end{figure}

{\underline{\congress}.~} Next, we want to explore scenarios where \emph{interactions between attributes} prove important to understanding properties of a graph.
To this end, we look at the \congress graph, where the two attribute vectors are binary vectors $\bv_d$ and $\bv_r$ corresponding to Democrat and Republican senators respectively (ignoring the small minority of independents).
We plot within-party agreement $(\bv_d^T \bS \bv_d + \bv_r^T \bS \bv_r)/2$ and cross-party agreement ($\bv_d^T \bS \bv_r$) over the years in Fig.~\ref{fig:polarization}.

We can observe that beginning from the 1990s, senators tend to agree among their parties, and disagree with the opposite party to a higher extent, hinting at a growing polarization in politics.
We note that agreement across parties is also low in 1937 (see the ``dip''), however, this is better explained by the fact that this congress had overwhelmingly more number of democrats. There is no hint of polarization for that instance, since there is no corresponding rise in the dashed (within-party) curve.
Fig.~\ref{fig:polarization} shows that aggregate functions from \method not only help us observe such phenomenon but also help quantify them to a relative extent.

{\underline{\states}.~} Lastly, we analyze \states, comparing within-state migration against cross-border migration for each of the 49 mainland states in the U.S.
We focus on {LDOS aggregate features}. this time using both positive and negative power functions,
 to analyze both short and long-range migration patterns. 
In other words, while small positive powers ($k$$=$$1$) capture local migration patterns, negative powers ($k$$=$$-1$) depicting paths of all lengths also reflect long-range migration behavior on a relatively global scale.

\begin{figure}[!ht]
		\vspace{-0.05in}
	\centering
	\begin{tabular}{cc}
		\includegraphics[width=0.47\linewidth]{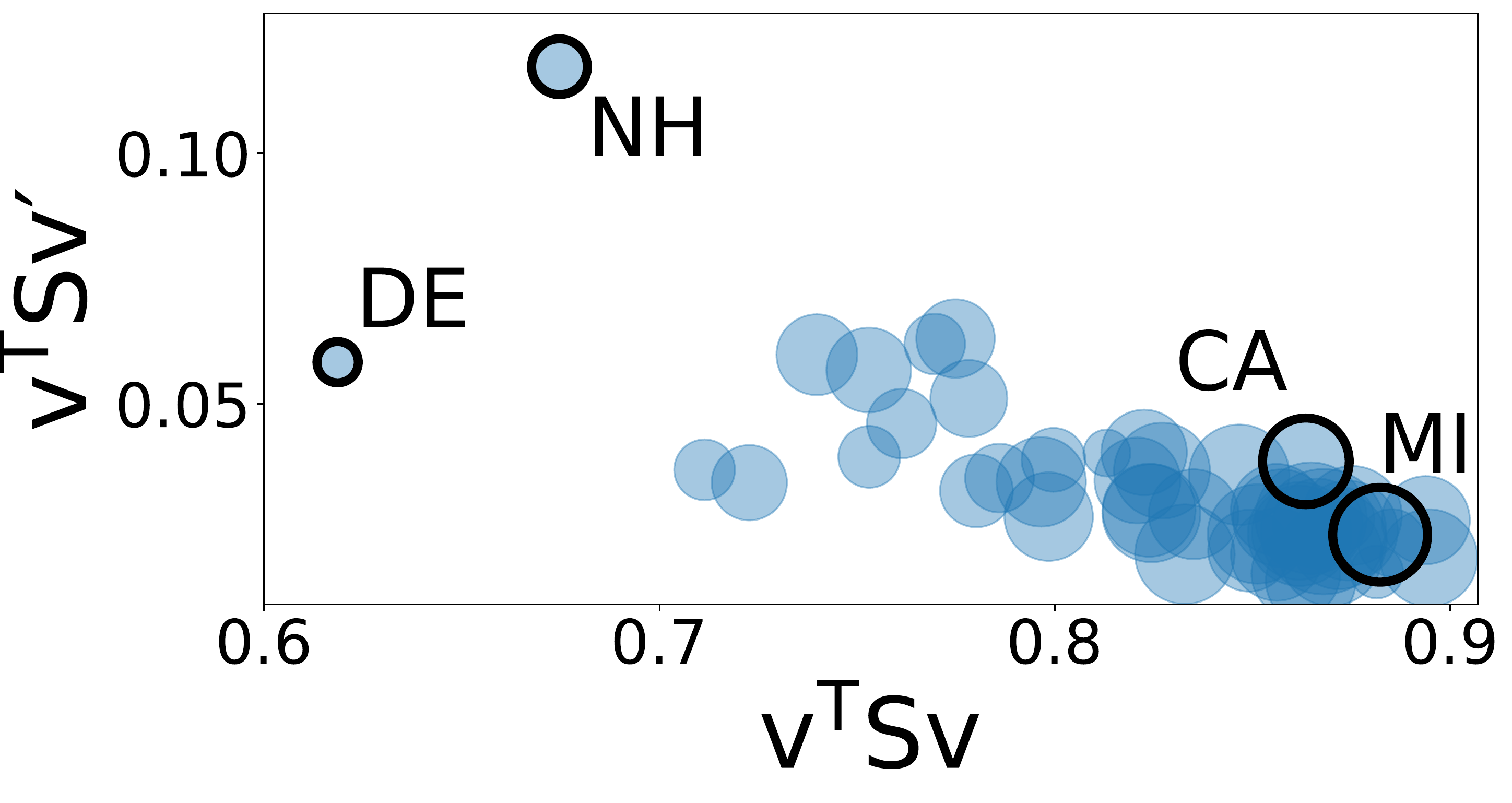}	& \includegraphics[width=0.47\linewidth]{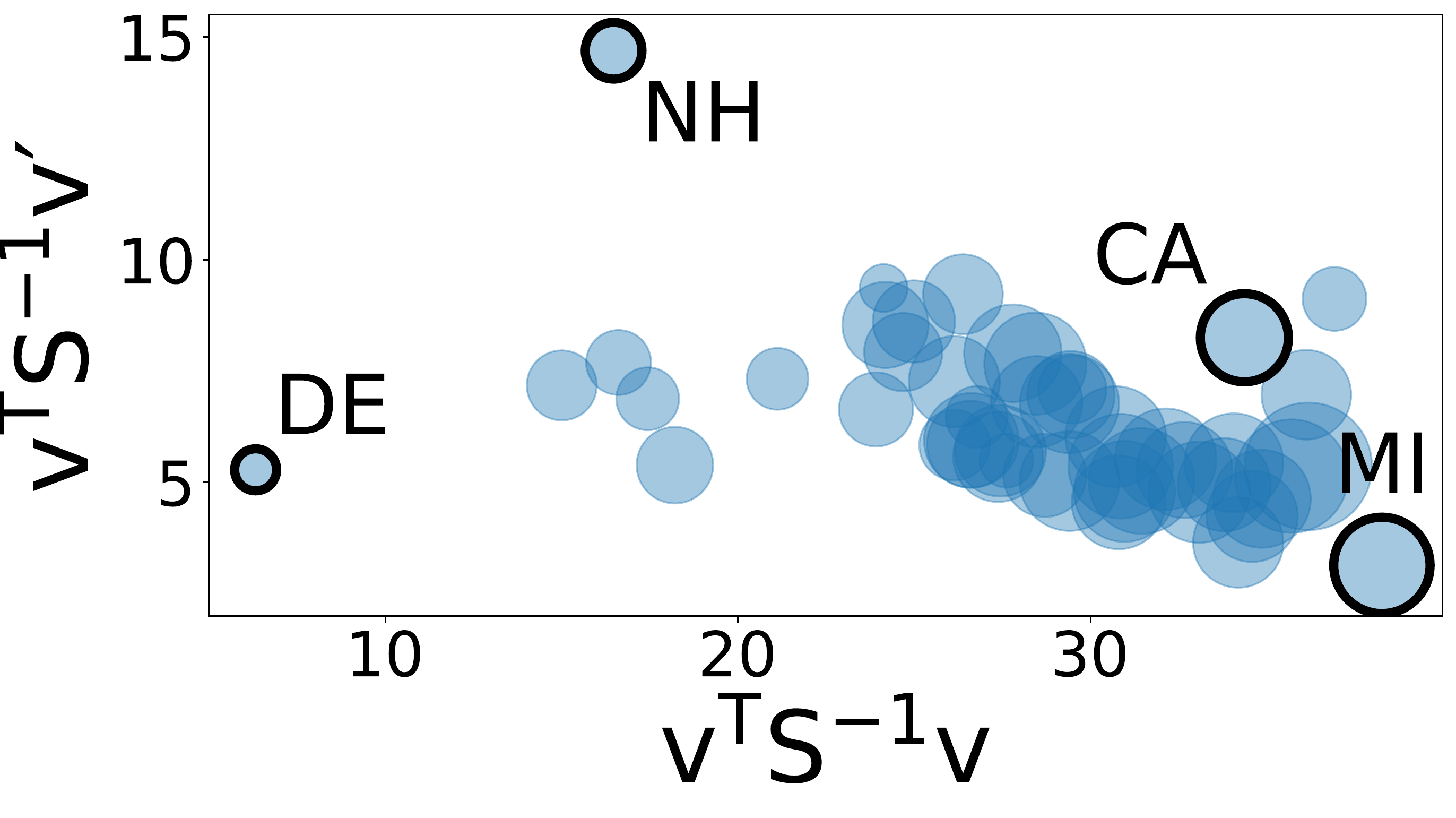} \\
	\end{tabular}
	\vspace{-0.1in}
	\caption{\label{fig:states_inout} Comparison of migration patterns for each U.S. state -- within its counties vs. across its borders; migration (left) over a local range, and (right) on a global scale. $\bv_w$ and $\bv_{b}$ refer to  binary vectors denoting within and border-state {counties}, respectively. Node sizes correlate to size of state.}
		\vspace{-0.05in}
\end{figure}

From Fig.~\ref{fig:states_inout}(left), we observe that at the local scale, most states have greater within-state migration than cross-border migration. Comparatively, NH and DE, being the states with the least number of counties, exhibit lower within-state migration.
Moreover, due to NH's geographical and political similarity with its bordering states, it shows highest cross-border migration.
On the other hand, larger states such as CA and MI exhibit mostly within-state migrations on the local scale. However, on the global scale (Fig.~\ref{fig:states_inout}(right)), the difference between these is more pronounced, since CA is a more popular long-range migration destination than MI.

%
%

\subsection{Scalability}

\method is not directly comparable to all the baselines in terms of
 resource requirements. 
 GNNs and \gvec need GPU processing, which make them incomparable to CPU-based \method and the rest.
Other differences, such as supervised training and collective processing of the graphs via multiple passes over the dataset (in contrast to one-by-one/independent processing by \method) put them in a different ``league''.
  
On the other hand, kernel baselines need considerably more memory. \wl, \wloa and \pk compute intermediate data (e.g. compressed labels) based on all the graphs in memory. These and \dos produce a $N$$\times$$N$ kernel matrix that is also memory-resident.

\begin{figure}[t]
\vspace{-0.1in}
	\centering
	\begin{tabular}{c}
		\includegraphics[width=0.6\linewidth]{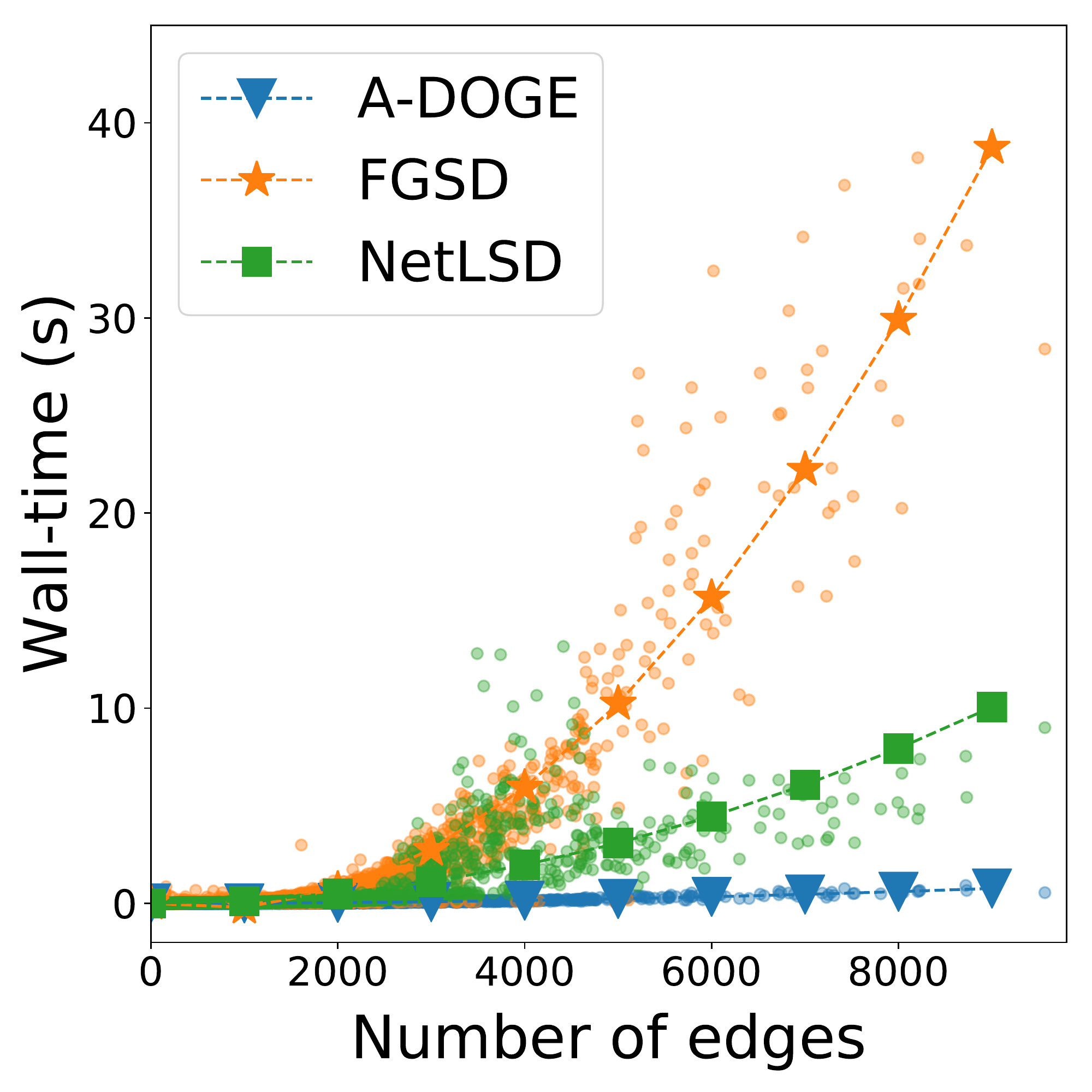} \\
	\end{tabular}
\vspace{-0.1in}
	\caption{Runtimes per graph in the \reddit dataset for each of the \method and the two baselines which can compute embeddings independently. \label{fig:time}}
	
\vspace{-0.3in}
\end{figure}

\fgsd and \netlsd are comparable in the sense that, similar to \method, they process the graphs independently one-by-one. Likewise, they are also unsupervised. However, they cannot handle node labels/attributes.
Nevertheless, we provide running time and scalability comparison in Fig. \ref{fig:time} that
plots the runtime vs. size in number of nodes for individual graphs in the \reddit dataset.
For any graph from the dataset (up to ~9500 edges), \method does not take more than 1 second to compute.
Fig.~\ref{fig:crown} compares this runtime for 3 of our largest datasets.
We can see that \method achieves the best time-accuracy trade-off among competing baselines.
For methods with comparable or better accuracy scores (e.g. \gin), \method is almost twice as fast on average.
For baselines with similar runtime (e.g., \wl), \method achieves significantly higher accuracy. 




\section{Conclusion}
\label{sec:conclusion}

We propose \method, an unsupervised graph embedding technique designed to efficiently capture structural properties as well as node labels and attributes of a graph. To this end \method uses spectral density, or density of states (DOS), derived from the eigenspectrum of the graph, as a tool to capture both global and local properties of a graph. 
Further, we extend local density of states to leverage node labels and attributes, and 
 capitalize on fast approximation algorithms making \method efficient and scalable to large graphs both in terms of time and space.
Being unsupervised, it is not only suitable for downstream supervised graph classification tasks, but also applies well to exploratory graph analysis. 
Through both  quantitative and qualitative experiments, we show the efficacy and efficiency of \method, where it outperforms unsupervised baselines and performs comparably to the supervised GNNs on graph classification tasks, and provides various insights into the analysis of real-world attributed graphs.




\section*{Acknowledgments}{
	\small{
	This work is sponsored by NSF CAREER 1452425. We also thank PwC Risk and Regulatory Services Innovation Center at CMU. Any conclusions expressed in this material are those of the authors and do not necessarily reflect the views of the funding parties.
}
}

\bibliographystyle{IEEEtran}

\bibliography{refs}

\end{document}